%% file: main.tex
\documentclass[Afour,sageh,times]{sagej}

\usepackage{moreverb,url}

\usepackage[colorlinks,bookmarksopen,bookmarksnumbered,citecolor=blue,urlcolor=blue]{hyperref}

\usepackage{tabularx}

\newcommand\BibTeX{{\rmfamily B\kern-.05em \textsc{i\kern-.025em b}\kern-.08em
T\kern-.1667em\lower.7ex\hbox{E}\kern-.125emX}}

\input{generated_values}

\begin{document}



\title{Bye-bye, Bluebook? Automating Legal Drudgery With AI-Augmented Rule Following}

\author{Matthew Dahl\affilnum{1} and Eric Mart\'inez\affilnum{2}}

\affiliation{\affilnum{1}Yale Law School, \href{mailto:matthew.dahl@yale.edu}{matthew.dahl@yale.edu} \\
\affilnum{2}Texas A\&M University School of Law, \href{mailto:ericmart@law.tamu.edu}{ericmart@law.tamu.edu}}

\begin{abstract}
One of the central promises of legal AI is to automate drudgery---the formal, repetitive tasks of lawyers' work that consume time without calling for much discretion. Yet it remains an open question how well AI models actually perform on such tasks. This article presents the first empirical examination of AI performance on perhaps the most ubiquitous and lamented form of legal drudgery: citation formatting under the \textit{Bluebook}. We make four contributions. First, we develop a new benchmark of \nTotal\ \textit{Bluebook} queries and show that, on average, frontier language models produce a fully compliant legal citation only \pooledAccuracy\ of the time in a zero-shot setting. Second, we conduct an experiment with five top law reviews and show that even a ``reasoning'' model falls far below the average score of the human candidates in these journals' annual editor-selection competitions. Third, we show that simply providing the models with the rules offers only modest improvements, calling into question the ability of retrieval-augmented generation (RAG) to ensure rule-following alone. Finally, we develop an approach that does meaningfully improve compliance: a neuro-symbolic system that first uses a model to parse natural language into structured citation elements, and then delegates the formatting to a deterministic rule-execution engine. This approach achieves an average accuracy increase of \sGain{} percentage points and total accuracy of up to \bestStructuredAccuracy{} on our benchmark. These results point toward a reorientation for legal AI. The original promise of automating drudgery still remains out of reach for even frontier language models on their own---but pairing them with symbolic rule engines may offer a tractable path forward.
\end{abstract}


\maketitle


\section{Introduction}
\label{sec:introduction}

In April 2025, an attorney at Latham \& Watkins---one of the world's preeminent law firms---needed help. She was preparing to file an expert declaration for a case in the Northern District of California, but she wanted to ensure that a supporting source was cited correctly.  So she did what a growing number of lawyers now do: she asked an artificial intelligence (AI) chatbot to format the citation for her.

\begin{figure}
    \centering
    \includegraphics[width=\columnwidth]{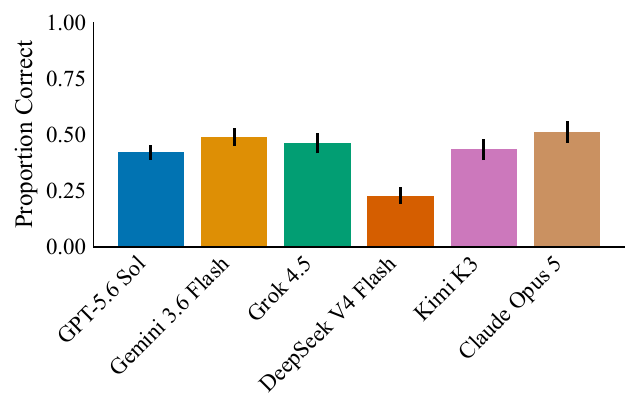}
    \caption{Flagship AI models achieve between \worstModelAccuracy{} (\worstModelName) and \bestModelAccuracy{} (\bestModelName) accuracy on \textit{Bluebook} formatting tasks ($n=\nPooled$ open tasks per model). Black bars represent 95\% confidence intervals.}
    \label{fig:results_by_model}
\end{figure}

There was just one problem. Although the attorney provided the chatbot with ``the correct publication title, publication year, and link to the provided source,'' the citation that the chatbot returned ``included an inaccurate title and incorrect authors.''\endnote{Declaration of Ivana Dukanovic Related to ECF No. 365, Concord Music Grp., Inc. v. Anthropic PBC, No. 24-CV-03811 (N.D. Cal. May 15, 2025), Dkt. No. 371 \P\ 6.}  Instead of simply rearranging the information the attorney had already verified, the system silently reinvented it, transforming an originally accurate submission into a plausible-looking, but ultimately false, doppelg\"anger of a citation. The attorney then filed this fictitious material with the court, affixing her expert's affirmation that the declaration's contents were true and correct ``under penalty of perjury.''\endnote{Declaration of Olivia Chen in Support of Anthropic's Sampling Proposal in Connection with Joint Discovery Dispute, Concord Music Grp., Inc. v. Anthropic PBC, No. 24-CV-03811 (N.D. Cal. Apr. 30, 2025), Dkt. No. 341-2 \P\ 23.}

Fabricated AI-generated citations are, unfortunately, no longer a novel phenomenon \citep{Charlotin2026, Liu2026b}. But the Latham attorney's mistake was not a ``hallucination'' in the conventional sense \citep{Dahl2024}. The attorney had completed the substantive research herself, identifying the source, verifying its contents, and supplying the relevant publication details. She asked AI only to perform the ministerial task of reformatting that information. In other words, she left the drudgery to the machine. And the machine failed at the drudgery.

This paper takes mistakes like these as its point of departure. One of the central promises of AI is that it will automate the repetitive, time-intensive, low-discretion work that consumes professional attention but demands relatively little professional judgment, thereby freeing workers to focus on higher-value tasks \citep{ranganathan_ye_2026_ai_intensifies}. This promise has been especially pronounced in the legal industry, where vendors market AI systems as tools for automating routine legal work, streamlining workflows, and enabling lawyers to concentrate on strategy, advocacy, and client service. Thomson Reuters, for example, promotes its legal AI offerings as helping legal professionals ``make quick work of routine tasks'' and focus on ``higher-level strategy,'' freeing attorneys to attend to the ``most rewarding'' aspects of their practice \citep{thomsonreuters_legal_ai_technology,thomsonreuters_2025_workflows}. Similarly, Harvey, a prominent legal tech startup, claims that its AI agents can ``execute legal work end to end, so you can focus on what only lawyers can do.'' Its tagline: ``Delegate the Work. Own the Judgment.'' \citep{Harvey2026}.

We investigate whether this division---between AI for drudgery and AI for judgment---is empirically defensible. So far, legal scholarship has focused overwhelmingly on examining the latter type of AI, from AI for legal research \citep{Afane2026, Magesh2025, Surani2026}, to legal analysis \citep{Blair-Stanek2023, Choi2024, Guha2024}, to legal drafting \citep{Chien2025, Schwarcz2026, Unikowsky2025}, to legal interpretation \citep{Arbel2024, Coan2025, Engel2024, Kruse2025, He2026, Martinez2026}, to even judicial decision-making and sentencing \citep{Kieffaber2025a, Liu2024, Posner2026a}. In other words, scholarly attention has gravitated toward evaluating precisely the forms of legal work that AI is ostensibly supposed to be \textit{leaving} to human professionals---once the drudgery had been automated.

By contrast, the more basic question of whether AI can reliably perform that legal drudgery in the first place has received far less attention. We know little, for example, about AI's ability to compute timing requirements from decentralized procedural triggers, follow the conditions of local rules and judges' standing orders, populate standardized legal templates and forms, cross-reference terms and concepts across documents, extract key facts and chronologies from legal corpora, screen filings for completeness against enumerated criteria, verify that cited authorities actually stand for the propositions attributed to them, or---as the Latham attorney found out the hard way---simply format legal citations in conformity with a publicly specified rule set. Even as legal vendors market their AI tools as being able to automate precisely these types of tedium, transparent benchmarks remain lacking \citep{Guha2026, Kapoor2024}, and the scholarly literature has been preoccupied with more ``open-textured'' \citep{Hart1994} applications instead.

This paper begins to fill that gap. We focus on perhaps the most paradigmatic form of legal drudgery: the formatting of citations in compliance with the \textit{Bluebook}, the definitive manual for citation in the American legal system \citep{ColumbiaLawReview2020}. Bluebooking is an unusually clean empirical target for several reasons. First, it is ubiquitous: citation formatting is a foundational, labor-intensive task performed by every law student, clerk, and practicing attorney. Second, it is formally specified: the \textit{Bluebook}'s 500+ pages of rules are publicly available, explicit, and, in principle, mechanically followable. Third, compliance is objective: unlike ill-defined judgments about legal meaning, the correctness of a citation can be verified by comparison to an authoritative answer key. And fourth, it can be high-stakes in practice---errors can result in waived arguments, judicial sanction, and, in the case of AI-fabricated citations, professional discipline.

Beyond serving as a test for legal drudgery, Bluebooking also offers a useful proxy for legal rule-following more generally. A foundational ambition of early legal AI scholarship was to render law computationally tractable by translating legal rules and procedures into algorithms that a machine could consistently apply \citep{buchanan1970speculation,sergot1986bna,governatori2022thirty,mccarty1977taxman}. And today, a growing theoretical literature in the AI-risk and alignment space contends that advanced AI systems must be designed to follow legal rules \citep{OKeefe2025,maas2025treaty,Nay2023}. If AI systems are indeed capable of internalizing and applying such rules, one would expect them to perform well in a domain like Bluebooking, where the governing instructions are formal, detailed, and publicly specified.

We show that they do not. In two different settings, we demonstrate that even frontier language models---including Claude Opus 5, ChatGPT 5.6 Sol, and Kimi K3---are unable to reliably follow the rules of the \textit{Bluebook}. First, we show that even when they are given fully accurate source information, just like the Latham attorney, they achieve only \worstModelAccuracy{}--\bestModelAccuracy{} accuracy on an original benchmark of \nTotal\ \textit{Bluebook} queries that we develop and make available as part of this paper. And second, in an experiment that we conducted in partnership with five T30 law schools, we show that the models consistently underperform humans in real law review write-on competitions.

We then turn our attention to exploring mechanisms for improving compliance. A common intuition is that retrieval-augmented generation (RAG) is a promising way to boost the performance of language models, yet we find only modest evidence for the efficacy of this technique in our setting. We therefore propose and develop an alternative architecture: a lightweight, neuro-symbolic system that first uses an AI model to parse natural language into structured citation elements, and then delegates the formatting to a deterministic rule-execution engine that we validate on our benchmark dataset. While simply giving the models direct access to the rules raises accuracy to only \icIncontextAccuracy{}, our AI-augmented rule pipeline achieves \bestStructuredAccuracy{}.

The remainder of this paper is organized as follows. Section~\ref{sec:bluebook_background} gives background on legal citation and the \textit{Bluebook}. Section~\ref{sec:study1} introduces our benchmark and presents our zero-shot evaluation of six frontier language models. Section~\ref{sec:study2} reports our pre-registered experiment with human law review write-on candidates. Section~\ref{sec:study3} demonstrates the inadequacy of RAG as a solution for the models' poor rule-following. Section~\ref{sec:study4} develops and evaluates our hybrid pipeline that pairs a model with a deterministic \textit{Bluebook} formatter, showing that this structured approach substantially improves compliance. Finally, Section~\ref{sec:discussion} discusses the implications. While automating legal drudgery still remains out of reach for language models on their own, pairing them with symbolic rule engines may offer a tractable path forward.

\begin{figure*}
    \centering
    \includegraphics[width=0.9\textwidth]{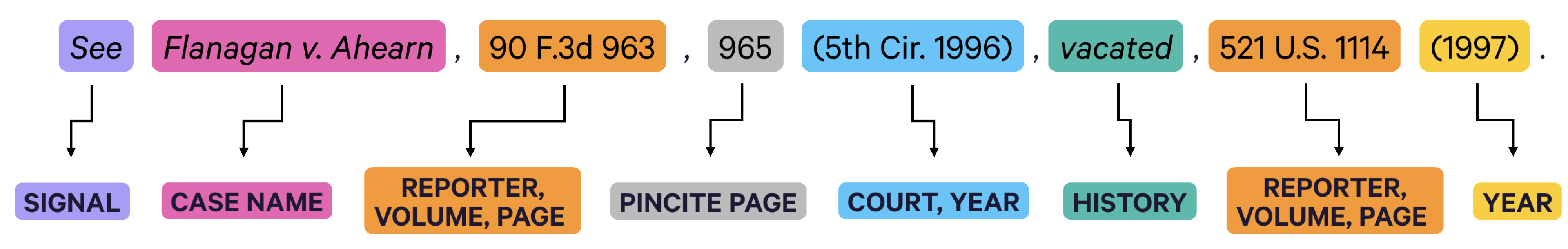}
    \vspace{0.1cm}
    \caption{Structure of a \textit{Bluebook}-compliant citation to case law. The \textbf{signal} indicates the weight that the citation is intended to convey. The \textbf{case name} reflects the names of the parties, which must be consolidated and abbreviated if there are many. The \textbf{reporter, volume, and page} give the physical location where the case is printed. The \textbf{pincite page} provides the specific page to which the citation points. The \textbf{court and year} give the jurisdiction and date of the issuing court. The procedural \textbf{history} notes any subsequent history of the case, such as if it has been vacated by an appellate court. The second \textbf{reporter, volume, and page} give the physical location where the appellate court's opinion is printed. The second \textbf{year} gives the date of that appellate opinion; no court is specified in this particular citation because the reporter for the appellate opinion (U.S.) unambiguously indicates that the appellate court is the Supreme Court.}
    \label{fig:citation_diagram}
\end{figure*}

\section{Background on the \textit{Bluebook}}
\label{sec:bluebook_background}

Legal citation is the backbone of legal practice. Far more than most disciplines, law is built on authority, meaning that the \textit{source} of a proposition can be a meaningful warrant for its acceptance \citep{Schauer2008}. Indeed, ``authority and hierarchy play a role in law that would be inimical to scientific inquiry'' \citep{Posner1990}. A properly formed citation can tell the reader ``at a glance'' whether the source is ``law at all'' \citep{Salmon2016}, what role it is serving in a given argument \citep{Gupta2025}, and what weight it should be accorded \citep{Dahl2024a}.

Today, the rules of American legal citation are organized in \textit{The Bluebook: A Uniform System of Citation} \citep{ColumbiaLawReview2020}. Known as simply the \textit{Bluebook} for short, this 500+ page manual dates back to the 1920s and describes, in great detail, the precise way in which almost every conceivable document is to be cited in official legal writing \citep{Shapiro2016}. For example, a full case law citation must contain any signal glossing the valence of the citation, the names of the parties, the page and volume of the reporter publishing the opinion, any pincite page, the court and year of the decision, and any subsequent history noting if the case was later vacated or abrogated. Abstruse styling rules for italicization, underlining, and the use of small caps must be respected as well. For instance, when the signal ``\textit{e.g.}" is used in conjunction with another signal, it must be preceded by an italicized comma but must be followed---for whatever reason---by an \textit{unitalicized} comma. Even more complicated rules govern the citation of statutes, legislative history, dictionaries, court filings, and numerous other categories of authority. We depict the basic format of a \textit{Bluebook}-compliant citation to case law in Figure~\ref{fig:citation_diagram}.

Over the years, the extraordinary complexity of the \textit{Bluebook} has garnered much criticism. Judge Richard Posner, in particular, has written a series of diatribes against the system \citep{Posner1986, Posner2011, Posner2016a}. He describes it as ``560 pages of rubbish, a terrible time waster for law clerks" and recommends that readers ``burn all copies" \citep{Posner2016a}. Others argue that ``\textit{Bluebook} skills simply signal membership in an elite club" \citep{Salmon2016}; that, ``[a]s a semiotic code, the \textit{Bluebook} effectively hides ideological roots and assumptions behind a mask of necessity and naturalness" \citep{Bacchus2002};  and that ``\textit{The Bluebook} hammers law students with the notion that the law is simply an intricate set of rules that, although tedious to learn, contains determinate answers decipherable by anyone willing to spend sufficient time staring at all of the possibly relevant rules" \citep{Heifetz1999}. At the same time, still others have spoken out in favor of the \textit{Bluebook}, arguing that it serves a useful pedagogical and information-retrieval purpose \citep{Asbury2011, Shawler1992, Whisner2008}.

In this paper, we are agnostic on this debate. For better or worse, the \textit{Bluebook} exists and is followed religiously by large numbers of law students, clerks, judges, and lawyers \citep{Heifetz1999}. A failure to adhere to its rules can lead to confusion \citep{Robbins1999} and even judicial sanction \citep{Fischer1997, Gallacher2007}. As the Seventh Circuit has remarked, ``[j]udges are not like pigs, hunting for truffles \dots\ [we will] strike any of the parties’ factual assertions, in any section of their briefs, that lack direct citation to easily identifiable support in the record.''\endnote{Gross v. Town of Cicero, Ill., 619 F.3d 697, 702 (7th Cir. 2010) (cleaned up).} Likewise, in another case, where ``[d]efendants' failure to provide a full date of publication prevented the Court from finding the cited authority in Westlaw,'' the court ``ADMONISHED [counsel] to provide pinpoint citations in future filings,'' stating that ``[i]n future rulings, the Court will DISREGARD citations to caselaw that are not supported by pinpoint citations.''\endnote{SASC, LLC v. Sch. Supply Connection, Inc., No. 3:23-CV-083, 2024 WL 3849424, at *8 (S.D. Ohio Aug. 15, 2024).} And, of course, the advent of AI has introduced a new dimension to this problem: citations to sources that do not even exist in the first place. As one court has explained, ``many harms flow from such deception---including wasting the opposing party's time and money, the Court’s time and resources, and reputational harms to the legal system (to name a few).''\endnote{Morgan v. Cmty. Against Violence, No. 23-CV-353, 2023 WL 6976510, at *8 (D.N.M. Oct. 23, 2023).} ``Courts therefore do not, and should not, make allowances for a party who cites to fake, nonexistent, misleading authorities.''\endnote{Kohls v. Ellison, No. 24-CV-3754, 2025 WL 66514, at *5 (D. Minn. Jan. 10, 2025) (cleaned up).}

Despite the foundational role that citation plays in legal practice, it is widely acknowledged that ``everybody hates bluebooking'' \citep{Franklin2008}. ``[L]ine editing for hours at a time with a copy of the infamous and opaque \textit{Bluebook} in hand,'' reports one former law review editor, is ``mindless scut work'' \citep{Oleson2004}. Yet it is precisely because Bluebooking is so irksome, we argue, that it is a promising candidate for automation. If AI could reliably follow the rules of the \textit{Bluebook}, then AI could replace a large amount of human labor currently performed by law students, clerks, and lawyers that is dedicated to completing drudgery of this type.

\begin{figure*}
    \centering
    \includegraphics[width=0.9\textwidth]{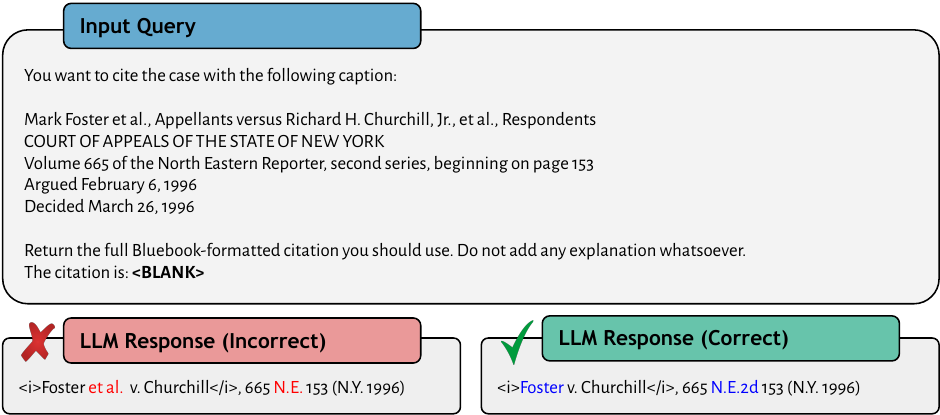}
    \vspace{0.05cm}
    \caption{Example query from the case law task. The model is provided with the caption to a case in natural language and is asked to complete the \textit{Bluebook} citation. Other tasks mimic this structure. Responses are graded against ground-truth answers sourced from \textit{Bluebook} experts.}
    \label{fig:example_task}
\end{figure*}

\section{LLMs Have Limited Internal Knowledge of the \textit{Bluebook}}
\label{sec:study1}

We begin by establishing a baseline of AI's inherent ability to implement the \textit{Bluebook}, without assistance from a human supervisor or a curated list of the applicable rules. To do so, we introduce the first \textit{Bluebook} benchmark: \nTotal\ gold-labeled queries that we carefully compile and make available alongside this paper.

We test performance separately for the two main sections of the twenty-first edition of the \textit{Bluebook}: the Bluepages and the Whitepages. The Bluepages govern the formatting of practitioner-facing legal documents (briefs, judicial opinions, motions),  while the Whitepages apply to academic legal writing and law review footnotes. These domains differ in important ways. Bluepages citations often involve a relatively stable set of primary authorities and standardized formats, whereas Whitepages citations are more heterogeneous, spanning a wider range of secondary sources and editorial conventions common in scholarship. Testing both regimes therefore helps distinguish whether AI can handle both a constrained, practice-oriented citation environment as well as the broader---and often more varied---citation landscape that dominates legal academia.

\subsection{Benchmark Construction}

Our main benchmark consists of \nPooled{} open-style citation-formatting queries. Each consists of entirely truthful and complete information about the authority---as might be retrieved from a document found in a legal research database like Westlaw or Lexis---and instructions about what kind of citation the model should construct. For example, Figure~\ref{fig:example_task} depicts the structure of a query asking for the creation of a case law citation. Specific examples for each query type are available in Appendix~\ref{sec:appendix_tasks}. For ease of exposition, we group them into three main categories:

\begin{enumerate}
    \item \textbf{Case law tasks.} Tasks related to constructing complete case citations from raw source information---encompassing party-name abbreviation, reporter information, court abbreviations and opinion dates, and, where the source calls for them, parallel reporters, subsequent case history, parentheticals, and short forms.
    \item \textbf{Enacted law tasks.} Tasks related to constitutional provisions; federal statutes (both annotated and unannotated); state statutes (both printed and electronic); statute short forms; and administrative law.
    \item \textbf{Other tasks.} Tasks related to other miscellaneous legal sources, including legislative resources (e.g., floor statements, committee reports, resolutions); court filings (e.g., depositions, appellate records, exhibits); and secondary sources (e.g., law review articles, books, newspapers, websites).
\end{enumerate}

The benchmark additionally contains \nClozeTotal{} cloze-style queries, in which the model is asked to fill in an isolated fragment---e.g., a reporter abbreviation, a court--date parenthetical, or an introductory signal---within an otherwise complete citation. Rather than testing end-to-end citation formatting, these cloze queries are helpful for diagnosing the models' specific failure points.

To build the Bluepages portion of the benchmark ($n=\nBluepagesPerModel$), we compiled and transformed questions from two legal writing textbooks: the \textit{Interactive Citation Workbook for The Bluebook: A Uniform System of Citation} \citep{Norton2024} and \textit{Understanding and Mastering The Bluebook: A Guide for Students and Practitioners} \citep{Barris2020}. Each contains a number of exercises that are explicitly designed for law students to practice their \textit{Bluebook} skills. The former is used in ``nearly 140" legal research and writing programs nationwide. Importantly, both contain expert-verified answers for each exercise, guaranteeing that we have a fully accurate set of gold labels against which to benchmark the AI models' performance.

To build the Whitepages portion ($n=\nWhitepagesPerModel$), we scraped the online versions of recently published Harvard Law Review (HLR) articles and randomly sampled footnotes as our raw material. We chose to use HLR footnotes for three reasons: (1) HLR citations are produced and checked by highly trained editors, so they are, in principle, especially likely to reflect correct \textit{Bluebook} form; (2) because the citations come from academic law review writing, they reflect the kinds of sources and citation patterns that dominate Whitepages usage in practice; and (3) HLR publishes fully formatted citations in HTML, closely mirroring the format of the Bluepages exercises and reducing transcription noise that could arise from manual conversion. After scraping these footnotes, we audited them against the governing \textit{Bluebook} rules for idiosyncrasies and made a small number of rule-verified corrections to them. More details on our scraping and cleaning pipeline are available in Appendix~\ref{sec:appendix_whitepages_dataset_details}.

\subsection{Experimental Setup}

We test both the Bluepages and Whitepages queries using six flagship, general-purpose large language models (LLMs). Each of these represents the latest models of their respective AI lab at the time of this study:

\begin{itemize}
    \item OpenAI's \texttt{\modelGpt} \citep{OpenAI2026a}
    \item Anthropic's \texttt{\modelClaude} \citep{Anthropic2026a}
    \item Google's \texttt{\modelGemini} \citep{Google2026}
    \item Moonshot AI's \texttt{\modelKimi} \citep{KimiTeam2026}
    \item DeepSeek's \texttt{\modelDeepseek} \citep{DeepSeek-AI2026}
    \item xAI's \texttt{\modelGrok} \citep{xAI2026a}
\end{itemize}

As our default---for reasons of efficiency and cost---we run these models in their standard (non-reasoning) mode. For robustness purposes, in a secondary analysis (Section~\ref{sec:reasoning_results}), we additionally evaluate \thinkingModelName{} in ``reasoning" mode and obtain convergent results.\endnote{We decided to run the ``reasoning'' experiment using \thinkingModelName{} because OpenAI's models remain the most popular models among the general public and thus are likely the models that users (whether lawyers, law students, or the general public) will be most likely to turn to for the purposes of generating a legal citation. Additionally, in our write-on experiment in Section~\ref{sec:study2} below, we also use an OpenAI model in reasoning mode, and for consistency we also run a similar model here.} We set the decoding temperature to 0 for all models.

To measure correctness, we use exact string matching between the model's response and the known \textit{Bluebook} answer. In particular, we penalize typeface deviations (e.g., small caps, italics, and other required typographic conventions) in addition to ordinary character deviations, given that typeface is an essential part of Bluebooking. In legal writing, students, lawyers, and editors are not permitted to ignore it simply because the intended referent is otherwise recoverable.\endnote{We also report in the Appendix an alternative, more generous scoring variant that does not penalize typeface deviations.} At the same time, we do not penalize models for superficial artifacts unrelated to \textit{Bluebook} compliance. We therefore do not treat minor formatting noise---such as trailing periods, incidental whitespace differences, or line breaks---as errors. We also apply light normalization to prevent unfair penalties stemming from typographic variants that are functionally identical in ordinary text environments (e.g., differing Unicode characters for hyphens and dashes, curly versus straight quotation marks and apostrophes, and similar punctuation variants). These preprocessing steps are applied symmetrically across model outputs and answer keys and are intended to ensure that errors reflect genuine \textit{Bluebook} mistakes rather than quirks of interface rendering or character encoding.

Our main estimand of interest is the models' population-level accuracy at open-style Bluebooking. Our reporting metrics therefore weight performance on the three task categories by how often the corresponding citation types actually appear in legal writing, rather than by their raw prevalence in our benchmark. Specifically, within each \textit{Bluebook} regime (Bluepages and Whitepages) we first compute a score for each task category (case law, enacted law, other) by averaging its tasks' accuracies, and then combine the three group scores using weights derived from the empirical distribution of citation types across \hlrN{} HLR footnote citations: case law \hlrWeightCases, enacted law \hlrWeightStatutes, and other sources \hlrWeightOther.\endnote{Each of the \hlrN{} HLR citations was assigned a fine-grained citation type during dataset construction. We map each type to the benchmark category whose tasks cover it, drop the \hlrUncoveredN{} citations (\hlrCoverageRate{} coverage) whose types no benchmark task covers---foreign and international materials and a long tail of rare source types (agency adjudications, interviews, working papers, and the like)---and renormalize the shares.} The headline pooled accuracy is the average of the resulting Bluepages and Whitepages scores.

\subsection{Benchmarking Results}

\begin{figure*}
    \centering
    \includegraphics[width=0.9\textwidth]{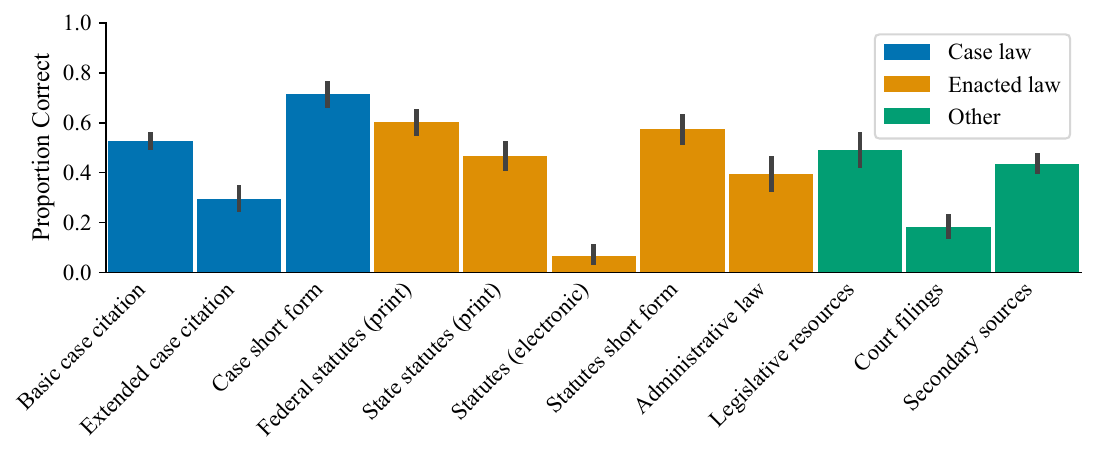}
    \vspace{0.1cm}
    \caption{Flagship AI models vary greatly in their ability to handle different \textit{Bluebook} assignments. The models perform comparatively well on case law citations (blue)---shown broken down into basic citations (full citations without a parenthetical or prior/subsequent history), extended citations (full citations with a parenthetical or prior/subsequent history), and short-form citations---and much worse on most enacted law (orange) and other (green) tasks (all models pooled). Black bars represent 95\% confidence intervals. Table~\ref{tab:overall_ti} in the Appendix reports the unpooled results for all models.}
    \label{fig:results_by_task_case_subcats}
\end{figure*}

Figure~\ref{fig:results_by_model} above reports overall accuracy for each of the six LLMs, pooled across the benchmark's open-style formatting tasks with the citation-frequency weighting described above. Across all tasks and models, performance is poor: the AI systems fail to produce a fully \textit{Bluebook}-compliant citation more than half the time (pooled accuracy = \pooledAccuracy). No model reaches even 55\% accuracy; \modelClaude{} leads at \claudeAccuracy{}, and \worstModelName{} trails at \worstModelAccuracy{}. Thus, our top-line takeaway is straightforward: even the latest state-of-the-art AI models cannot be used as substitutes for competent \textit{Bluebook} diligence. Naively using these models for citation generation produces error rates that would be unacceptable in professional or scholarly legal writing.

At the same time, performance is highly heterogeneous across task types. Figure~\ref{fig:results_by_task_case_subcats} disaggregates accuracy at the task level, and we turn to discussing these more granular results next.

\subsubsection{LLMs Perform Best at Case Law Formatting, but Still Unreliably}

Of the three task categories, the models perform comparatively well on case law---but even here accuracy falls well short of what Bluebooking demands (Figure~\ref{fig:results_by_task_case_subcats}). The models do best on short-form citations (\fullCaseShortFormAccuracy), where a pre-existing citation is given and they only need to reshape it into a pincite or \textit{id.} citation. They do worse on formatting basic citations (\fullCaseBasicAccuracy{}), where they need to create a full citation from scratch. Accuracy falls sharpest for extended citations (\fullCaseExtendedAccuracy{})---those that add a parenthetical (stating, for instance, if the cited opinion is a dissent, a concurrence, \textit{en banc}, etc.) or append prior/subsequent case history to a basic citation.

Using our cloze tasks to decompose the case law results further, we observe similar results (Figure~\ref{fig:results_by_task_cloze}). Models are middling on the mechanical core of a case citation---reporter abbreviations (\caseReportersAccuracy) and court--date parentheticals (\courtDateAccuracy)---and weaker on procedural detail such as subsequent history (\caseHistoryAccuracy) and parentheticals (\caseParentheticalsAccuracy).

\subsubsection{LLMs Struggle with Statutes, Regulations, and Secondary Sources}

Moving beyond case law, performance is best---but still poor---on printed federal statutes (\fedStatutesAccuracy). Printed state statutes, which are more complex, are worse (\stateStatutesAccuracy). And mirroring the short form case law results, the models perform relatively well on the short form statute queries (\statuteShortFormAccuracy).

The failures become more pronounced for sources that require more specific tailoring. For administrative law citations (e.g., to the Code of Federal Regulations or the Federal Register), pooled accuracy is \adminLawAccuracy. For legislative resources (e.g., committee reports, floor statements, and resolutions), pooled accuracy is \legResourcesAccuracy. For secondary sources (books, journal articles, newspapers, dictionaries), models manage only \secondarySourcesAccuracy{} accuracy. For court filings (e.g., depositions, appellate records, exhibits), accuracy drops to \courtFilingsAccuracy. And for electronic statutes, accuracy is close to zero: \elecStatutesAccuracy. In other words, once the task moves away from a relatively standardized citation template, models more often than not fail to produce \textit{Bluebook}-compliant outputs.

One plausible explanation for these sharp declines is that enacted law and secondary-source formats are structurally more heterogeneous than case law. Different states use distinct statutory publication conventions, and the same federal provision may appear in different authoritative or commercial compilations (e.g., the U.S.C., U.S.C.S., or U.S.C.A.), each triggering its own \textit{Bluebook} standard. Regulations likewise appear in multiple compilations (e.g., the C.F.R. and state analogues), again with source-dependent rules. And the electronic statute format is particularly arcane, requiring many abbreviations referring to the date and legislative session through which the electronic source is current.\endnote{For example: \texttt{Me. Stat. tit. 17, § 2512 (Lexis+ through Ch. 559 of 2024 2d Reg. Sess. of 131st Me. Leg.)}.} In these domains, even small differences in publisher, format, or currentness information change the required citation form, and the models frequently fail to track those contingencies.

\begin{figure}
    \centering
    \includegraphics[width=\columnwidth]{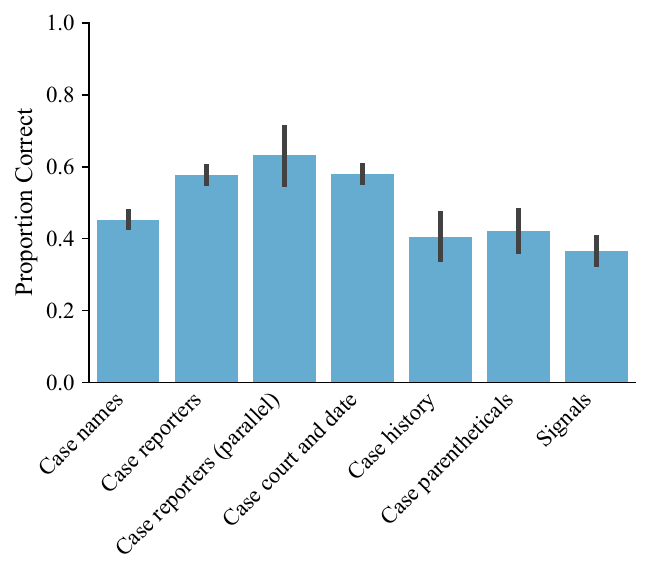}
    \caption{Flagship AI models struggle with cloze-style tasks where they are asked only to fill in a missing part of an otherwise correct \textit{Bluebook} citation (all models pooled). Black bars represent 95\% confidence intervals.}
    \label{fig:results_by_task_cloze}
\end{figure}

Finally, we also note that the AI models struggle with producing signals, achieving only \signalsAccuracy{} accuracy on this cloze-style task (Figure~\ref{fig:results_by_task_cloze}). To be sure, signals require more than mechanical formatting: they encode the relationship between authority and proposition, and even human writers sometimes disagree about the correct signal to use \citep{Robbins1999}. But the models' poor performance on even this cloze task reinforces the takeaway that when \textit{Bluebook} compliance depends on context-sensitive legal judgment (or on highly source-specific procedural detail), AI outputs are unreliable.

\subsubsection{LLMs Perform Much Worse on Whitepages Than on Bluepages}
\label{sec:whitepages_bluepages_gap}

\begin{figure}
    \centering
    \includegraphics[width=\columnwidth]{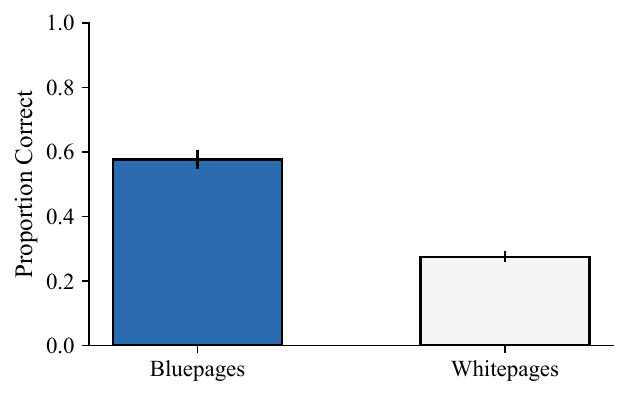}
    \caption{Flagship AI models perform substantially worse on Whitepages tasks (drawn from real-world law review footnotes) than Bluepages tasks (drawn from legal writing textbooks) (open tasks only, all models pooled). Black bars represent 95\% confidence intervals.}
    \label{fig:bluepages_vs_whitepages}
\end{figure}

The AI models also perform worse on the Whitepages tasks than on the Bluepages tasks. Pooled across models, mean accuracy is \whitepagesAccuracy{} on Whitepages compared to \bluepagesAccuracy{} on Bluepages---a gap of \bwGap{} percentage points (Figure~\ref{fig:bluepages_vs_whitepages}).\endnote{At the model level, the Whitepages drop is also universal. \modelGpt{} falls from \gptAccuracyBP{} accuracy on Bluepages to \gptAccuracyWP{} on Whitepages; \modelKimi{} from \kimiAccuracyBP{} to \kimiAccuracyWP; \modelGrok{} from \grokAccuracyBP{} to \grokAccuracyWP; \modelGemini{} from \geminiAccuracyBP{} to \geminiAccuracyWP; and \modelDeepseek{} from \deepseekAccuracyBP{} to \deepseekAccuracyWP. \modelClaude{} shows the smallest proportional drop (\claudeAccuracyBP{} to \claudeAccuracyWP) but still loses nearly twenty percentage points.}

This difference is especially striking because the prompts explicitly specify the target regime: for the Whitepages tasks, the models are told to produce citations suitable for a law review article, and for the Bluepages tasks, the models are told to produce citations suitable for professional documents such as briefs or memoranda. In other words, the Whitepages deficit is not simply a failure to guess which rule set applies---the models exhibit difficulty in reliably executing Whitepages conventions even when the relevant context is clearly provided.

One might initially suspect that the Whitepages deficit is driven largely by interface constraints---most notably, the difficulty of producing required typeface (especially small caps) in ordinary chat outputs. But our evaluation design makes that explanation unlikely. We query the models via their APIs and instruct them to express typeface using XML tags, thereby bypassing the limitations of plain-text rendering in chat interfaces.\endnote{For example, the models were instructed to indicate small caps with a \texttt{<span class="small-caps">…</span>} wrapper. If anything, this feature of our setup makes our estimates charitable. Real-world users typically do not post-process AI outputs with XML tagging conventions; they copy text directly into Word processors, briefs, or manuscripts and rely on the model to get formatting right (or at least flag where formatting is required). Allowing XML tagging thus reduces friction relative to ordinary usage, underscoring that the observed Whitepages failures are not an artifact of our evaluation medium but a genuine limitation in their \textit{Bluebook} rule-following abilities.} The Whitepages drop therefore cannot be explained as a purely technical inability to display typography. Rather, it reflects a deeper problem: models often fail to apply typeface correctly even when given a straightforward, machine-readable way to do so.

\subsubsection{``Reasoning" Does Not Meaningfully Improve Performance}
\label{sec:reasoning_results}

Another natural hypothesis is that LLMs' low zero-shot accuracy is a consequence of a lack of explicit ``reasoning"---test-time compute usage \citep{Snell2024}---and that a model with a reasoning mode enabled might perform significantly better.\endnote{Despite the anthropomorphic label, AI ``reasoning'' is simply the generation of intermediate tokens before returning a final answer to the user \citep{Kambhampati2026}.} However, while reasoning models have proven adept at certain legal tasks \citep{Schwarcz2026}, other research suggests that this mode can also cause catastrophic failures \citep{Kwon2026, Shojaee2025, Song2026b, Zhang2026}. Therefore, to test whether reasoning can improve legal rule-following specifically, we run our \textit{Bluebook} benchmark on a commonly used model (\thinkingModelName{}) in medium-thinking mode for comparison.

We find that enabling reasoning improves accuracy only modestly: \thinkingModelName{} achieves \gptNonthinkingAccuracy{} accuracy in non-thinking mode versus \gptThinkingAccuracy{} in medium-thinking mode (see Figure~\ref{fig:gpt56_ladder} below). This gain, while real, underscores the robustness of our earlier results: even with reasoning enabled, the poor performance we observe across models persists at levels far too high for reliable professional use.

Disaggregating the results by Bluepages versus Whitepages shows that the improvement is limited under both regimes. On the Bluepages tasks, accuracy rises from \gptNonthinkingAccuracyBP{} in non-thinking mode to \gptThinkingAccuracyBP{} in medium-thinking mode. On the Whitepages tasks, the relative improvement is larger---from \gptNonthinkingAccuracyWP{} to \gptThinkingAccuracyWP---but from a far lower base, leaving Whitepages accuracy at barely one in four even with reasoning enabled. We also note that these gains seem to be driven mostly by better typographic execution, especially on the Whitepages. If we ignore typeface in our scoring, the Whitepages accuracy increases only to \gptThinkingAccuracyWPNI{}, meaning that a large share of \thinkingModelName{}'s remaining Whitepages errors are still substantive, even in reasoning mode. These results suggest that test-time compute alone is unlikely to translate into reliable procedural compliance in domains where success hinges on both substantive and presentational conventions: a five-point gain still leaves the model failing more than half of all queries.

\subsubsection{LLMs' Mistakes Can Be Serious}

\begin{table*}[!t]
    \centering
    \begin{tabularx}{\textwidth}{XXX}
        \toprule
        \textbf{\textit{Bluebook} citation} & \textbf{LLM citation} & \textbf{Error} \\
        \midrule
        \textcolor{blue}{\textit{Campbell v. Ins. Serv. Agency}} & \textcolor{red}{\textit{Robert v. Campbell}} & Confuses and flips the parties \\
        \midrule
        \textit{Roland v. O'Neal}, 122 S.W. 827 (\textcolor{blue}{Ky.} 1909) & \textit{Roland v. O'Neal}, 122 S.W. 827 (\textcolor{red}{Ky. Ct. App.} 1909) & Misses that prior to 1976, the Court of Appeals was the highest state court in Kentucky \\
        \midrule
        \textit{Hyde v. Martin}, \textcolor{blue}{2017-CA-00822-SCT} (Miss. 2019) & \textit{Hyde v. Martin}, \textcolor{red}{283 So. 3d 1199} (Miss. 2019) & Invents completely non-existent citation for unpublished case \\
        \midrule
        \textit{Dunbar v. State}, 46 So. 3d 81 (Fla. Dist. Ct. App. 2010), \textcolor{blue}{\textit{rev'd on other grounds}}, 89 So. 3d 901 (Fla. 2012) & \textit{Dunbar v. State}, 46 So. 3d 81 (Fla. Dist. Ct. App. 2010), \textcolor{red}{\textit{rev'd}}, 89 So. 3d 901 (Fla. 2012) & Falsely suggests case is inapposite by misstating circumstances of reversal \\
        \midrule
        N.Y. Penal Law § 125.25(1) (\textcolor{blue}{McKinney 2020}) & N.Y. Penal Law § 125.25(1) (\textcolor{red}{Consol. 2020}) & Uses ambiguous and misleading publisher abbreviation \\
        \midrule
        2 U.S.C.\textcolor{blue}{A.} §§ 275-276 (\textcolor{blue}{West} 2016) & 2 U.S.C. §§ 275-276 (2016) & Fails to include attribution to annotated version of the code \\
        \midrule
        \textcolor{blue}{Chris Costain}, Note, \textit{The Automation of Transportation: The Advent of Autonomous Driving Technology}, 55 New Eng. L. Rev. 92 (2021) & \textcolor{red}{Note}, \textit{The Automation of Transportation: The Advent of Autonomous Driving Technology}, 55 New Eng. L. Rev. 92 (2021) & Erroneously suppresses identity of the author \\
        \bottomrule
    \end{tabularx}
    \vspace{0.3cm}
    \caption{Selected examples of substantive errors occurring in LLM-generated citations. Errors of this magnitude are rare overall, but serious when they occur. Consistent with the methodology depicted in Figure~\ref{fig:example_task}, in all cases the LLM is provided with enough raw material to generate the desired \textit{Bluebook} citation.}
    \label{tab:error_examples}
\end{table*}

Finally, it is useful to understand the specific nature of the models' \textit{Bluebook} mistakes in more depth, both quantitatively and qualitatively. When LLMs do make errors, how serious are those errors?

\begin{figure}
    \centering
    \includegraphics[width=\columnwidth]{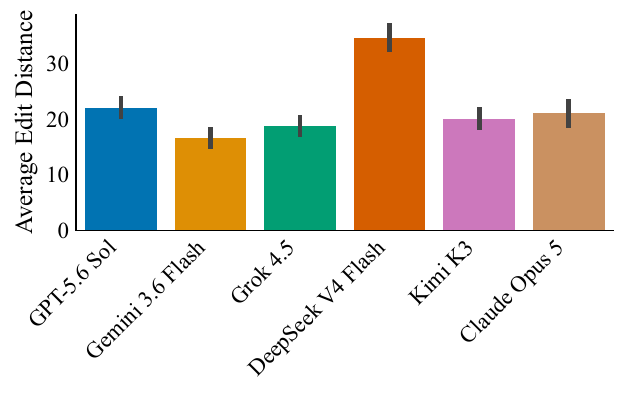}
    \caption{Flagship AI models require non-trivial numbers of edits to their incorrect responses to bring them into full \textit{Bluebook} compliance (open tasks only). Black bars represent 95\% confidence intervals.}
    \label{fig:edit_distance}
\end{figure}

Figure~\ref{fig:edit_distance} provides a numerical answer. For each incorrect response, we compute the Levenshtein edit distance: the minimum number of single-character insertions, deletions, and substitutions that are required to transform the model output into the ground-truth answer. Pooled across tasks, the mean distance among incorrect outputs is large, and the best-performing model by accuracy (\modelClaude) is still \claudeEditDist{} edits away from the correct answer on average.\endnote{If we ignore typeface when scoring, the mean edit distance falls to \pooledEditDistNI{}.} A similar pattern appears when we separate Bluepages from Whitepages (Appendix Figure~\ref{fig:edit_distance_by_page}). These metrics suggest that when models generate noncompliant citations, the problem is rarely a missing comma or stray hyphen; rather, the typical incorrect output requires dozens of character-level changes to fully conform to the \textit{Bluebook}.

Table~\ref{tab:error_examples} provides qualitative texture by cataloging representative failure modes across tasks. These examples confirm that the models' errors are not merely cosmetic. They include mistakes such as misstating the parties, supplying an incorrect court or date, misreporting the volume/reporter/page location, omitting or fabricating subsequent history, misidentifying a publisher or compilation (e.g., U.S.C. vs. U.S.C.A.), or misattributing an author. Each of these errors undermines the basic procedural function of citation: enabling verification, efficient retrieval, and accurate attribution.

To be clear, substantive failures of this magnitude are not the modal case. For many incorrect citations, a reader could still infer the intended source---especially when the core locator information is intact. But comprehensibility is not the standard that we are benchmarking. The \textit{Bluebook} is not designed to produce citations that are merely intelligible; that goal could be achieved with a far shorter and more forgiving manual \citep{Posner1986}. Rather, Bluebooking is an exercise in hyper-detailed rule-following compliance. On that metric, even seemingly minor deviations matter. (And the edit distance results show that model errors are frequently not minor, particularly in the Whitepages regime.)

\section{AI Is Worse Than Human Bluebookers}
\label{sec:study2}

We turn next to evaluating AI's capacity for Bluebooking in a more practice-proximate setting: the Bluebooking component of real law review ``write-on" competitions. The benchmark that we introduced in Section~\ref{sec:study1} serves as a useful baseline for assessing whether LLMs can follow detailed formatting rules at all, but it also invites a potential critique: perhaps its tasks are too artificial, too unlike how \textit{Bluebook} competence is assessed in practice, or too disconnected from the institutional settings in which citation errors actually matter.

Our analysis in this section addresses that set of concerns by using a human benchmark instead. The law review write-on exam is a canonical, high-stakes test of citation competence: it is administered under competition rules designed to differentiate applicants; it is graded by the very editorial boards that enforce citation norms; and (at many schools) it plays a meaningful role in determining admission to a flagship law review \citep{Volokh2010}. For our purposes, it is therefore a paradigmatic---and to our knowledge the most ecologically valid---benchmark of Bluebooking ability. Rather than asking whether AI models can perfectly solve stylized \textit{Bluebook} exercises, we now ask whether they can at least surpass humans in a real-world application.

\subsection{Experimental Design}

\subsubsection{Sampling Protocol}

To conduct this experiment, in the summer of 2025 we solicited participation from the top 30 flagship law reviews (as ranked by U.S. News \& World Report), following a research design and analysis plan that we pre-registered.\endnote{Our pre-registration is available at \url{https://aspredicted.org/tn4mm5.pdf}.} In total, six law reviews agreed to participate, shared their Bluebooking materials, and in principle agreed to allow us to use an AI model to take their exam, which the journal would then grade in a blind format as part of their normal process.\endnote{For two schools, the grading was done in an unblinded manner, as the grading for the competition had already concluded by the time of the study.} After we produced and transcribed model responses for all participating schools, one law review did not ultimately provide scoring information despite our attempts to follow up, leaving a final sample of five law reviews with usable outcome data. We anonymize the identities of all the schools in this paper to preserve the integrity of their selection processes.

\subsubsection{Prompting, Transcription, and Grading}

For each participating law review, we used the exact exam materials that it provides to its student candidates for its \textit{Bluebook} exam (e.g., citation questions, source packets, and any instructions or rubrics provided to applicants). We did not redesign questions, harmonize across schools, or convert the exams into lab-style templates. (Our aim was to preserve each exam's idiosyncrasies---its source mix, formatting requirements, and grading conventions---because those features are part of what makes the write-on exam an ecologically valid measure of Bluebooking competence.)

To give the AI test-taker the strongest opportunity for success, we pre-registered that we would use the most advanced ``reasoning'' model that was available: at the time, OpenAI's o3 model \citep{Openai2024b}. However, while we did end up using o3 for all but one of the exams, during the study, OpenAI released GPT-5 \citep{Singh2025a} and retired o3, forcing us to use that model (in reasoning mode) instead. Aside from this forced substitution on a single exam, we followed our pre-registration protocol.

We prompted the model in approximately the same way across all exams. In each case, we provided the exam's exact Bluebooking instructions and materials and instructed the model to respond as an examinee: follow the exam's directions, produce the required citations or edits, and avoid adding commentary. Because write-on exams vary in structure (e.g., whether they require edits to a source document, standalone citation strings, or \textit{Bluebook} corrections within prose), we allowed minor prompt adjustments as needed to ensure the model's output aligned with the exam's required response format. For the longer edit-style exams, we also broke the task into stages---first asking the model to correct an initial subset of footnotes, then the next subset, and so on---rather than asking it to edit the entire document in a single pass. The purpose of this batching was to give the model the best opportunity to complete the exam accurately, instead of setting it up for failure by overloading a single prompt. The goal throughout was to give models the best shot at the exam while staying within the confines of the exam's own submission rules.

To facilitate grading under ecologically valid conditions, we manually transcribed the model responses into the exact format required by each competition (most commonly, a Word document with changes tracked). This step was mechanical: it ensured that the model's answers were presented in the same medium and layout as human submissions, enabling grading to proceed using ordinary law-review workflows.

Whenever feasible, participating law reviews graded the model's transcribed submission blindly, as part of their normal grading process alongside human submissions, using the same rubric and procedures applied to applicants. When blind grading was infeasible in practice, the law review graded the model's submission unblinded but using the same scoring rubric and point allocations used for human exams. In all cases, our aim was to match real-world write-on grading practices as closely as possible.

\begin{table*}[!t]
    \centering
    \begin{tabular}{lcccc}
        \toprule
        \textbf{School} & \textbf{Number of Applicants} & \textbf{Write-On Cutoff}\endnote{In line with our pre-registration, we calculated this cutoff either (1) directly via precise information that was shared from the law review; or (2) indirectly based on a combination of the rank of the model and the number of candidates who typically make it onto the law review at the school. In particular, at Schools 1, 2 and 3, we were provided with the number of applicants, the full distribution of scores, and AI rank, and we inferred the rank cutoff based on the number of students who make law review for a given year. At School 4, we were provided with the number of applicants, AI rank, and rank cutoff directly. At School 5, we were provided with the full distribution of scores of those who graded on to law review, along with the AI's score in comparison to that distribution.} & \textbf{AI Rank} & \textbf{AI Percentile Among Accepted Scores} \\
        \midrule
        1 & 291 & 58th  & 131st        & 0th \\
        2 & 157 & 55th  & 133rd        & 0th \\
        3 & 119 & 60th  & 119th        & 0th \\
        4 & 85  & 45th  & 44th         & 2nd \\
        5 & --- & 32nd  & $>$ 32nd\endnote{At School 5, we were provided only with the distribution of scores among those who successfully wrote on, not the full applicant pool. The model's score fell below all successful candidates, so its precise rank among all applicants is unknown but exceeds the 32nd position.}     & 0th \\
        \bottomrule
    \end{tabular}
    \vspace{0.3cm}
    \caption{Breakdown of write-on results by school. The \textbf{Write-On Cutoff} column gives the rank of the lowest student who was actually admitted to the law review. For example, for School 1, the top 58 students successfully wrote-on. For comparison, the \textbf{AI Rank} column gives the rank of the AI model (for School 1, the AI placed 131st). Finally, the \textbf{AI Percentile Among Accepted Scores} column gives the percentile placement of the AI model among the students who were actually accepted to the law review.}
    \label{tab:exam_results}
\end{table*}

\subsubsection{Outcome Measures and Analysis}

For each exam---conditional on the participating law review's consent and data availability---we recorded: (1) the model's raw score; (2) the model's percentile relative to human test takers; (3) the number of human test takers; and (4) the approximate selection threshold implied by the number of applicants admitted to the law review based in part on write-on performance. Using these data, we were able to summarize model performance descriptively across exams (mean/median/min/max raw score and percentile) and calculate how often the model would have cleared the relevant admission threshold under each participating law review's process. To ease our analysis, we assumed that the \textit{Bluebook} exam was the only criterion for selecting law review editors. (Effectively, this amounts to assuming that the model would not have performed better or worse than the real students on any additional write-on components, such as a personal essay or their first-year grades.)\endnote{This simplifying assumption is conservative in the model's favor. Many journals incorporate additional selection components (e.g., personal statements, substantive editing exercises, grades, or other criteria), so limiting the analysis to Bluebooking alone therefore likely overstates the model's actual probability of ``writing on." Conversely, because law review membership is not determined solely by citation performance, the resulting percentiles should not be read as a direct comparison to the entire population of law review editors---though it is plausible that, within many journals, the subset of editors who do the most intensive cite-checking are disproportionately those who performed well on the Bluebooking component.}

\subsection{Write-on Experiment Results}

\subsubsection{AI Falls Far Short of Human Performance on Write-on Exams}

\begin{figure}
    \centering
    \includegraphics[width=\columnwidth]{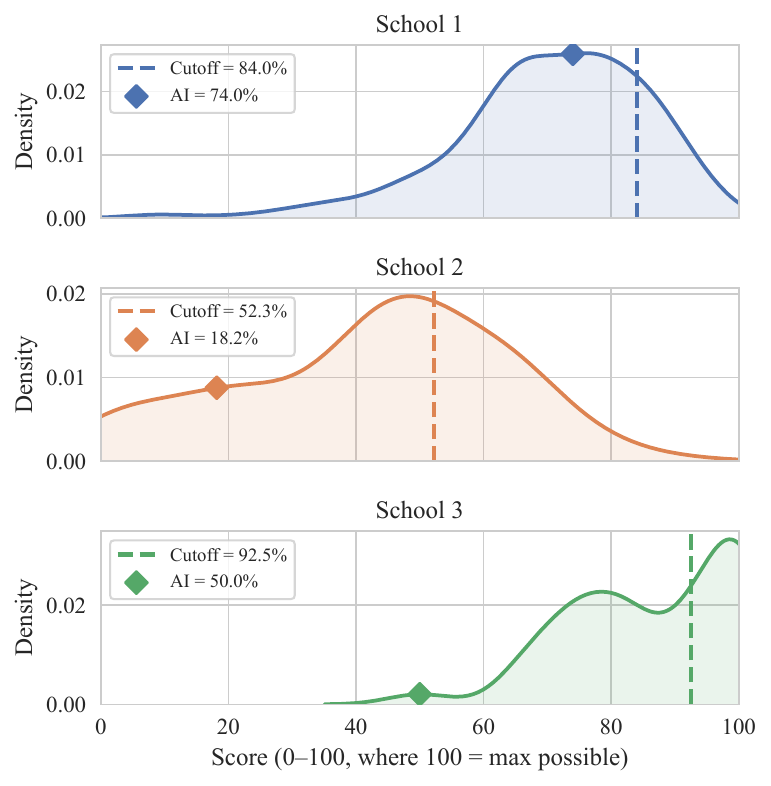}
    \caption{AI consistently underperforms students on law review write-on examinations. Only schools from which we received full distribution data are shown. Scores are normalized to a 0--100 scale for ease of comparison. Dashed lines indicate the write-on cutoff; diamond markers indicate the AI model's score.}
    \label{fig:schools_density}
\end{figure}

Table~\ref{tab:exam_results} reports the results of our experiment, showing that across the five participating law reviews, the AI model does not meet the performance threshold required to write-on. In every school for which we have scoring information, the model's overall score falls below the performance typical of successful write-on candidates. Indeed, across schools, the model's score is well below the average score of successful candidates. For the schools for which we have the full distribution of student results, Figure~\ref{fig:schools_density} depicts the AI model's performance relative to the students' graphically.

The model's relative placement among successful candidates is also illuminating. In four of the five schools, the model's score is below the score of every successful write-on candidate, implying that it would not have been admitted under the ordinary cutoff or ranking procedures. In the remaining school, the model's score exceeds only one successful candidate. Moreover, that school applies an additional policy under which applicants scoring below a specified Bluebooking threshold are not admitted absent remedial training; the model scores well below that threshold. Thus, under that school's own rules, the model would not be admitted as a regular write-on entrant.

\subsubsection{AI Struggles Most With ``Below-the-line" Tasks}

These results imply that AI cannot currently be used to replace human \textit{Bluebook} editors. Yet, if anything, our results likely overstate AI's real-world Bluebooking ability because write-on exams are not pure Bluebooking tests: many include substantial non-Bluebooking components (e.g., typographical corrections, proofreading, or other editorial tasks) that appear ``above-the-line'' in addition to the ``below-the-line'' footnote editing that is the focus of the \textit{Bluebook}. On the exams that emphasize below-the-line tasks, the model performs substantially worse, but on exams with more above-the-line elements, the AI model tends to perform comparatively well, as it is able to accrue points that are only indirectly related to \textit{Bluebook} mastery. Consequently, the model's overall exam score should be understood as only an upper bound on its true Bluebooking ability.\endnote{Qualitatively, this deficit appears to stem from a difficulty in reliably detecting and operating on footnote content: the model sometimes fails to notice that a citation appeared in a footnote, omits required edits to footnote material, or becomes more prone to hallucinating content when the relevant text is embedded below the line.}

A related source of errors that we observe concerns typeface, which is a persistent feature of below-the-line formatting in law reviews. Across all the exams, the model is particularly poor at producing italics and small caps. Because this typography is pervasive in law review citation form, this limitation is not merely cosmetic: under real grading practices, it mechanically converts responses that are otherwise substantively correct into point losses. Put differently, even when a model knows the right answer in plain text, it may still fail the write-on task because it cannot execute the required typographic transformations in the medium in which the exam is taken.

\section{Rule Retrieval Does Not Reliably Ensure Rule-Following}
\label{sec:study3}

So far, we have shown that off-the-shelf LLMs are unreliable at Bluebooking, as measured on two complementary benchmarks. The first is an \textit{absolute} benchmark: whether models can accurately produce compliant citations at a minimally acceptable rate (Section~\ref{sec:study1}). The second is a \textit{relative} benchmark: whether models perform well enough to satisfy the standards that legal institutions actually impose (Section~\ref{sec:study2}). In both settings, we show that they do not.

We now explore ways that the models' rule-following abilities might be enhanced. While techniques like instruction-tuning \citep{Colombo2024, Niklaus2025} and other forms of fine-tuning \citep{Blair-Stanek2024, Dominguez-Olmedo2025} are well-suited to applications where the rule set is relatively limited, we are interested in a setting where the rules are many and are not naturally available in a paired question-and-answer format. For this type of rule-following, researchers typically reach toward retrieval-augmented generation (RAG) as a solution \citep{Lewis2020}.

The core idea is that while it may be too much to ask LLMs to intuit rules from their training data alone, they might be able to perform well if they are directly prompted with the rules instead. Indeed, a recurring theme in the long-context literature is that many AI failures are driven less by ``reasoning" in the abstract than by missing or inaccessible information \citep{Li2024c, Wu2026, Zhang2024b}. Accordingly, modern workflows often focus on supplying the relevant text so the model can apply an explicit manual, specification, or policy to the task at hand.

To test whether rule access is the bottleneck, we examine whether a frontier long-context AI model can improve on our benchmark when it is provided with the applicable portions of the \textit{Bluebook} directly. If it can, then Bluebooking should become close to a solved problem once the relevant rules are placed in the prompt. If not, it suggests a deeper limitation: that even when models have direct access to the governing rules, they struggle to (i) identify the controlling provisions, (ii) synthesize them into an operational procedure, and (iii) execute that procedure with the level of exactness required by legal practice.

\subsection{Retrieval Methodology}

We perform this experiment using \modelGpt{}, the same model we evaluated in reasoning mode in the baseline benchmark in Section~\ref{sec:reasoning_results} and whose predecessor we evaluated in the write-on experiment in Section~\ref{sec:study2}. (ChatGPT is also the model that ordinary users are most likely to reach for when generating a citation.) We run the model in its standard (non-reasoning) mode as our main analysis: as Section~\ref{sec:reasoning_results} showed, enabling reasoning lifts the model's zero-shot accuracy only modestly, so the standard mode is the natural baseline for isolating the effect of supplying the rules.\endnote{For robustness purposes, we repeat the experiment with reasoning enabled and find a similarly modest increment over the non-reasoning in-context condition; Appendix Figure~\ref{fig:gpt56_ladder_thinking} reports the comparison.}

For each query, we provide the model with (i) the source information in the same format as in Section~\ref{sec:study1} and (ii) native PDF excerpts of the \textit{Bluebook} containing the rules and well-formatted examples relevant to the task. For example, case law tasks are accompanied by the pertinent portions of both the Bluepages and Whitepages Rules 2, 3, 4, 6, 7, 8, 10 and Tables 1, 6, and 8, \textit{inter alia}.\endnote{For reference, Table~\ref{tab:in_context_rule_list} lists the \textit{Bluebook} rules that were provided to the model for each task. The length of the supplied rule text varies between approximately 17,000 and 67,000 tokens depending on the specific task, making the prompt comparable in scale to other dense legal procedures that require navigating cross-references and executing detailed, exception-ridden requirements.} Effectively, we perform the retrieval step of a typical RAG setup ourselves, once again giving the model the best chance at success by removing any errors that might arise in that process.

Apart from supplying the relevant rules, the remainder of our methodology mirrors that of our initial benchmark to permit direct comparison between the results of that setup and the gains attributable to the in-context, rule-supplied setup. We decode at temperature 0 and score outputs using the same metrics as before: primary accuracy is assessed via exact string matching to the expert-provided ground truth, treating typeface as part of \textit{Bluebook} compliance (with light normalization for inconsequential whitespace and Unicode punctuation variants). To assess whether providing the rules significantly improves citation accuracy---and whether any effect varies by citation regime---we fit a logistic regression with binary correctness as the dependent variable and the in-context condition (zero-shot vs.\ rule-supplied), page type, their interaction, and task category as fixed effects, with standard errors clustered by query.

\subsection{Retrieval Results}

\begin{figure}
    \centering
    \includegraphics[width=\columnwidth]{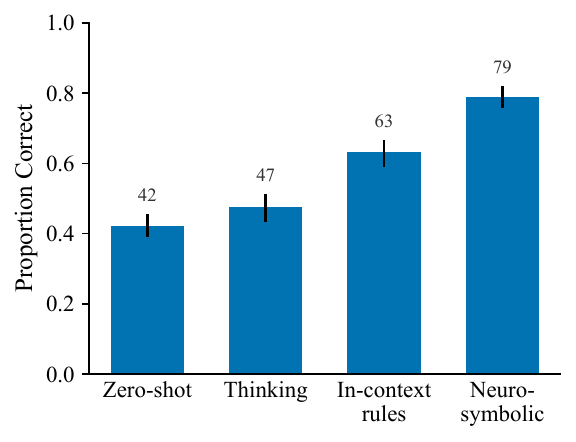}
    \caption{The four approaches tested in this paper, applied to a single model (\modelGpt): zero-shot prompting, reasoning mode, in-context rule provision, and the neuro-symbolic (AI-augmented) pipeline. The neuro-symbolic architecture yields by far the largest improvement, roughly doubling zero-shot accuracy. The in-context and neuro-symbolic conditions are run without reasoning (Appendix Figure~\ref{fig:gpt56_ladder_thinking} reports the results with reasoning enabled). Black bars represent 95\% confidence intervals.}
    \label{fig:gpt56_ladder}
\end{figure}

We find that providing \modelGpt{} with the task-relevant \textit{Bluebook} rule excerpts helps---but still falls well short of solving the problem. Pooled across the open query set, accuracy rises from \icZeroshotAccuracy{} in the zero-shot condition to \icIncontextAccuracy{} in the in-context condition, a gain of \icGain{} percentage points (Figure~\ref{fig:gpt56_ladder}, \bbP). The gain appears in both citation regimes: Bluepages accuracy rises from \icZeroshotAccuracyBP{} to \icIncontextAccuracyBP{}, and Whitepages accuracy from \icZeroshotAccuracyWP{} to \icIncontextAccuracyWP{} (\bbInterP).\endnote{Under typeface-insensitive scoring, the in-context coefficient remains positive and significant ($\beta$ = \bbBetaNI, \bbPNI), while the page-type interaction reverses sign ($\beta$ = \bbInterBetaNI, \bbInterPNI): the disproportionate Whitepages gain under our primary metric is driven largely by improved typeface execution when the model can see the rules' formatted examples.} Yet even with the governing rules in its prompt, the model still fails more than a third of all tasks, and over half of the Whitepages tasks---error rates that remain unacceptable in professional legal writing.

At the task level, the effects are highly heterogeneous. The largest gains come from tasks where the model appears able to apply the supplied rule text directly: legislative resources jump from \icLegresourcesZs{} to \icLegresourcesIc, secondary sources from \icSecondarysourcesZs{} to \icSecondarysourcesIc, state statutes from \icStatestatutesZs{} to \icStatestatutesIc, federal statutes from \icFedstatutesZs{} to \icFedstatutesIc, and basic case citations from \icFullcasebasicZs{} to \icFullcasebasicIc. But other tasks barely move even with the governing rules in hand: electronic statutes stay flat at \icElecstatutesIc{} and administrative law (\icAdminlawZs{} to \icAdminlawIc) and the statute short form (\icStatuteshortformZs{} to \icStatuteshortformIc) improve only marginally. The page-type breakdown is revealing as well: court filings, for instance, improve on the Bluepages (from \icCourtfilingsZsBP{} to \icCourtfilingsIcBP) but remain stuck at \icCourtfilingsIcWP{} on the Whitepages. Where the citation format is most arcane, in other words, rule access alone does not produce rule execution. We report full per-task results in Appendix Figure~\ref{fig:in_context_results} and in Appendix Table~\ref{tab:in_context_bp_wp}.

Overall, these findings complicate the prevailing wisdom that RAG is a reliable route to legal rule-following. Supplying the governing rules directly does help---consistent with RAG's successes in other legal settings \citep{Nay2024, Schwarcz2026}---but the ceiling is relatively low. Even with the retrieval problem solved by construction and the exact controlling provisions placed in the prompt, the model still cannot execute the rules with the exactness that legal practice demands.

\section{AI-Augmented Rule Following}
\label{sec:study4}

If supplying the \textit{Bluebook} directly to an LLM offers only modest improvements, then the bottleneck may not be access to the rules but the execution of them. This points to an alternative strategy. Rather than asking a model to apply the rules itself, we can delegate that step to a system that applies them deterministically. In other words, we propose an \textit{AI-augmented} approach to Bluebooking: a model parses, and a rule-execution engine formats.

This type of hybrid approach to legal automation echoes the original ambition of legal AI. Early work in the field sought to render well-specified legal rules computationally tractable by encoding them as executable logic \citep{buchanan1970speculation, mccarty1977taxman, sergot1986bna}, but that project was limited by the difficulty of formalizing rules by hand and mapping unstructured inputs onto them. LLMs, by contrast, are plausibly well-suited to precisely those front-end tasks, even as they remain unreliable at the exact rule application that a symbolic engine can perform trivially. A growing strand of literature accordingly pairs the two, using an LLM to first structure natural-language material and then delegating inference to a deterministic system \citep{Janatian2023, Jurayj2026, Kant2025, Sunny2026, Yadamsuren2025}.

We propose taking this neuro-symbolic approach to Bluebooking. Instead of asking an LLM to do free-form citation generation, we decompose the task into two components. The neural component---a standard LLM---handles the parts of the problem that require language understanding: identifying the citation type, extracting the relevant bibliographic elements from the user's query, and mapping them to a structured representation. The symbolic component---a deterministic engine---handles the parts of the problem that require precise, reproducible rule application: taking the structured data as input and computing the correctly formatted citation string by executing the encoded \textit{Bluebook} logic exactly.

This design separates two failure modes that we hypothesize are driving the poor performance that we have observed in the foregoing sections of the paper. The first is \textit{extraction failure}: the model fails to correctly identify or parse the relevant citation elements from the input. The second is \textit{rule-following failure}: the model correctly identifies the elements but applies the \textit{Bluebook} rules incorrectly. By routing all rule application through our deterministic engine, we eliminate the second failure mode entirely for citation types that the engine is built to handle.

\subsection{Algorithm Design}

\subsubsection{A Deterministic \textit{Bluebook} Formatter}

The first component of our pipeline is a deterministic \textit{Bluebook} formatter that takes a structured record describing a legal source and returns a fully formatted citation. While a small number of proprietary legal citation tools exist \citep{Bennett2026, CaseBriefly2026, LegalEase2026}, none to our knowledge have been validated against a public test bank, nor have they published accuracy results on a comprehensive task suite. This is unsurprising: validation requires a benchmark, and until the present work, no comprehensive benchmark of \textit{Bluebook} tasks with expert-verified answers had been assembled.

We designed our formatter $\widehat{\text{R}}$ with the goal of approximating, as close as possible, the true \textit{Bluebook} rule set $\text{R}$, implementing its rules as code rather than natural-language \citep{Martinez2026a}. To validate that our formatter $\widehat{\text{R}}$ is indeed close to $\text{R}$, we first tested it on our original benchmark queries from Section~\ref{sec:study1}, but converted into structured JSON records with one field per citation element. For example, a query like ``You want to cite a case with the name Susette Kelo, et al., Petitioners v. City of New London, et al., a 2005 United States Supreme Court case reported in volume 545, page 469, of United States Reports'' becomes the following structured record:
\begin{quote}
\footnotesize\ttfamily
\{\\
\hspace*{1em}"citation\_type": "case",\\
\hspace*{1em}"case\_name\_full": "Susette Kelo,\\
\hspace*{2em}et al., Petitioners v. City of\\
\hspace*{2em}New London, et al.",\\
\hspace*{1em}"volume": 545, "page": "469",\\
\hspace*{1em}"reporter\_full": "United States\\
\hspace*{2em}Reports",\\
\hspace*{1em}"court\_full": "Supreme Court of\\
\hspace*{2em}the United States",\\
\hspace*{1em}"year": 2005\\
\},
\end{quote}

\noindent which the formatter then renders as \textit{Kelo v. City of New London}, 545 U.S.\ 469 (2005). On the full structured task suite, the formatter alone---with no LLM in the loop---achieves \formatterCeilingAccuracy{} accuracy (Bluepages: \formatterCeilingBP; Whitepages: \formatterCeilingWP). This perfect performance allows us to infer that whatever errors arise downstream are attributable to the natural-language parsing step, not to the rule-execution step.

To ensure that our formatter is not simply overfitting to our initial test suite, we also constructed an entirely new set of queries drawn from HLR footnotes, with no overlap with the original benchmark ($n$ = \formatterHlrN). Running on this held-out set, the formatter achieves \formatterHlrAccuracy{} accuracy, confirming that its ceiling-level performance on the original suite reflects genuine rule coverage and generalizability.

\subsubsection{The AI-Augmented Pipeline}
\label{sec:ai_augmented_pipeline}

\begin{figure}
    \centering
    \includegraphics[width=0.7\columnwidth]{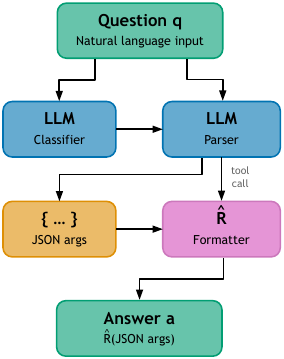}
    \vspace{0.05cm}
    \caption{Overview of our AI-augmented pipeline. LLMs (blue nodes) first classify the natural-language query into a citation type and parse its content into a standardized JSON schema. We then feed this structured data into our formatter $\widehat{\text{R}}$ (purple node), a deterministic engine that applies the \textit{Bluebook} rules.}
    \label{fig:pipeline_diagram}
\end{figure}

The second component of our pipeline is an LLM that converts a natural-language query into a structured record of the kind shown above. While existing regular expression tools are adept at parsing citations that already appear in a recognizable form \citep{Cushman2021}, only a language model can flexibly transform unstructured text into a standardized format. We prompt the model to do this natural language conversion in two phases. First, we ask the model to \textit{classify} the query into the most likely citation type (e.g., case law, statute, court document, etc.) based on a list of heuristics. Second, we ask the model to \textit{parse} the query content into structured JSON. Depending on the classification result in the first phase, we dynamically inject the appropriate JSON schema and a set of 2--9 few-shot examples into the context window to maximize the model's ability to complete this task.\endnote{More technically, we implement structured generation via the providers' tool-use (function-calling) interface \citep{anthropic2024tool_use, openai2024structured_outputs}. Tool use is supported by all six providers we test in the models' standard (non-reasoning) mode. The one exception arises in the reasoning-mode variant of the pipeline reported in Appendix Figure~\ref{fig:gpt56_ladder_thinking}: OpenAI's API rejects tool calls when reasoning is enabled on \modelGpt{}, so that run uses a JSON-prompt fallback that asks the model for a JSON object matching the same schema and parses the response. Additional implementation details, including the exact tool schemas and prompts, are provided in Appendix~\ref{sec:appendix_study4}.}

Crucially, the LLM is not asked to format anything itself. The structured output that it produces is fed into our deterministic formatter $\widehat{\text{R}}$, which actually applies the rules and renders the formatted citation. An illustrative example of the full pipeline---natural-language query, then LLM-produced JSON, then formatter-rendered citation---is depicted in Figure~\ref{fig:pipeline_diagram}.

We test the pipeline end-to-end using the same six frontier models as before: \modelClaude{}, \modelGpt{}, \modelGemini{}, \modelDeepseek{}, \modelGrok{}, and \modelKimi. We score the resulting citations using the same procedure as well: exact string matching to the expert-provided ground truth, with light Unicode normalization, and treating typeface as part of \textit{Bluebook} compliance. To assess whether the structured pipeline significantly improves citation accuracy and whether the effect varies by regime or model, we fit a logistic regression with binary correctness as the dependent variable and condition (zero-shot vs.\ structured-pipeline), page type, their interaction, task category, and a model fixed effect as explanatory variables, with standard errors clustered by query.

\subsection{Results}

\begin{figure*}
    \centering
    \includegraphics[width=0.9\textwidth]{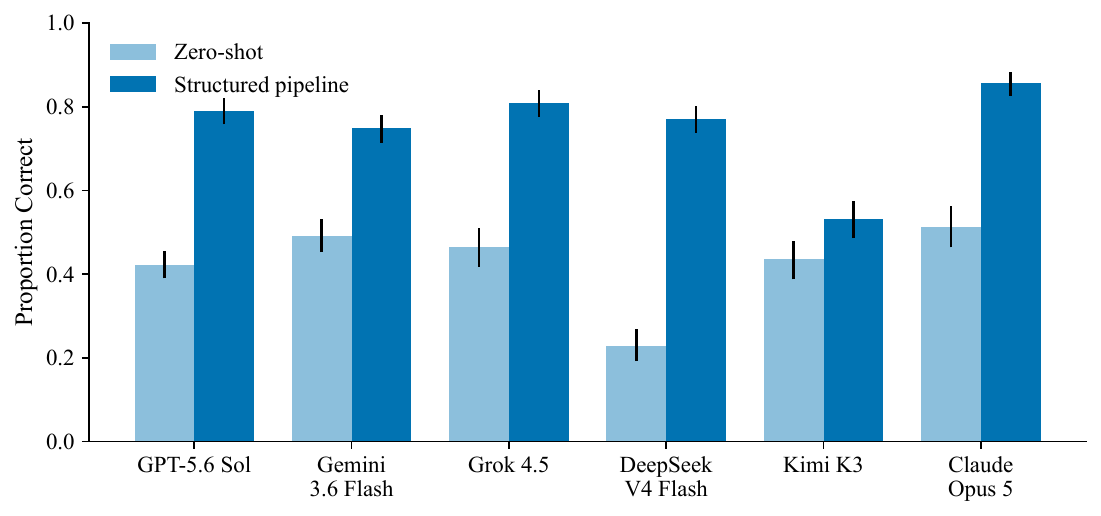}
    \vspace{0.1cm}
    \caption{The AI-augmented pipeline substantially outperforms zero-shot prompting for every model in our lineup. Light bars represent zero-shot accuracy; dark bars represent accuracy under the structured pipeline combining an LLM parser with a deterministic formatter (all open tasks pooled). Black bars represent 95\% confidence intervals.}
    \label{fig:structured_by_model}
\end{figure*}

Pooled across all six models, accuracy increases from \sZeroshotPooled{} in the zero-shot condition to \sStructuredPooled{} in the structured-pipeline condition---a gain of \sGain{} percentage points (\sP) that holds for every model in our lineup, for every task category, and for virtually every individual task (Figure~\ref{fig:structured_by_model}). The contrast with the RAG approach is notable: whereas simply giving the model the rules in its prompt does not meaningfully improve accuracy, pairing the model with a deterministic rule-execution engine does. Figure~\ref{fig:gpt56_ladder} makes the comparison clear by comparing the four approaches covered in this paper across a single model (\modelGpt): enabling ``reasoning'' improves the zero-shot baseline of \ladZeroshot{} only modestly (to \ladThinking); providing the rules in-context helps more (\ladInContext) but still leaves over a third of tasks failing; and the neuro-symbolic architecture yields by far the largest improvement, nearly doubling zero-shot accuracy.

\subsubsection{Gains Hold Across Tasks}

Looking at each task individually, the largest improvements appear precisely where we previously observed the most acute zero-shot deficits (Appendix Figure~\ref{fig:structured_results_by_task}). Electronic statutes, which the LLMs alone could complete correctly only \elecStatutesAccuracy{} of the time, rise to \sElecStatutesSt{} under the structured pipeline. State statutes (print) climb from \sStateStatutesZs{} to \sStateStatutesSt; legislative resources from \sLegResourcesZs{} to \sLegResourcesSt; secondary sources from \sSecondarySourcesZs{} to \sSecondarySourcesSt; and court filings from \sCourtFilingsZs{} to \sCourtFilingsSt. The pattern is consistent across the case law category as well, with basic case citations rising from \sFullCaseBasicZs{} to \sFullCaseBasicSt{} and extended case citations from \sFullCaseExtendedZs{} to \sFullCaseExtendedSt. In short, the architecture does not merely shift the easy tasks higher; it improves every task in the benchmark, largely closing the gap across the board.

\subsubsection{Gains Hold Across Models}

Every model in our lineup outperforms its own zero-shot baseline (Figure~\ref{fig:structured_by_model}).\endnote{\modelKimi{}'s gain is the smallest in the lineup. Much of the shortfall is provider-side: \modelKimi{}'s API host supports tool calling, but unlike the other providers we test it does not validate the returned tool-call arguments against the supplied schema, so its extractions more often contain malformed or misplaced fields that the deterministic formatter cannot render. Its structured-pipeline accuracy should therefore be read as a lower bound on what the same model would achieve behind a schema-enforcing endpoint.} Notably, the model that is weakest in the zero-shot condition (\modelDeepseek, at \sDeepseekZs) actually surpasses the performance of the strongest zero-shot model (\modelClaude, at \sClaudeZs) once it is augmented with the structured pipeline (rising to \sDeepseekSt). This is what one would expect if the binding constraint on Bluebooking accuracy is rule execution rather than parsing capability: once the rule-execution step is delegated to a deterministic engine, model differences in raw \textit{Bluebook} knowledge stop mattering as much, and the differentiator is each model's linguistic ability.

\subsubsection{Gains Hold Across Bluepages and Whitepages}

The structured pipeline also substantially narrows the Bluepages--Whitepages gap that our initial benchmark revealed (\bwGap{} percentage points pooled across models). On Bluepages tasks, pooled accuracy rises from \sZeroshotPooledBP{} to \sStructuredPooledBP; on Whitepages tasks, from \sZeroshotPooledWP{} to \sStructuredPooledWP. The Whitepages gain (\sGainWP{} percentage points) is larger than the Bluepages gain (\sGainBP{} percentage points), consistent with our previous finding that the Whitepages deficit was driven in substantial part by typeface errors that the deterministic formatter renders correctly by construction.\endnote{Under typeface-insensitive scoring, the same pattern holds, with somewhat smaller gains because there are no typeface errors inflating the original gap.}

\subsection{Two Advantages of an AI-Augmented Architecture}

Our results suggest that the AI-augmented pipeline enjoys two advantages over unstructured approaches to legal AI integration. The first is a reduction in drudgery that does not even require an LLM in the loop. Going from a structured representation of source elements to a fully formatted \textit{Bluebook} citation is, on its own, an enormous time savings for any lawyer, clerk, or law-review editor who already has the source information in hand---which, in many practical workflows, is the modal case. This is why proprietary citation web apps already exist; what our results add is a validated, comprehensive accuracy baseline (\formatterCeilingAccuracy) that the industry has lacked. The second, more ambitious, advantage arises from routing the natural-language query through an LLM parser and then through the formatter. This yields a system that ingests the same kind of free-form input that proved challenging in the zero-shot context, yet produces \textit{Bluebook}-compliant output at rates that approach human-editor reliability on many tasks. Both advantages flow from the same architectural principle---separating parsing from rule execution---and both depend on the formatter actually working on structured input, which our validation confirms.

\section{Discussion}
\label{sec:discussion}

Can AI automate legal citation formatting? In the first comprehensive empirical examination of AI Bluebooking, we show that the answer is still no---at least not without additional structure and engineering. These findings carry several implications. First, they call into question the implicit assumption that the automation of relatively routine, low-discretionary tasks is already a solved problem (Section~\ref{sec:discussion_drudgery}). Second, they complicate broader aspirations in the AI governance space for the development of ``law-following'' AI (Section~\ref{sec:discussing_law_following}). And third, they suggest and provide a blueprint for a reorientation of legal AI scholarship: away from a preoccupation with what H.L.A. Hart famously called ``open-textured'' tasks \citep{Hart1994} and toward what we tentatively label a ``formal turn'' in reliability instead (Section~\ref{sec:discussion_formal_turn}).

\subsection{Bluebooking and the Unfulfilled Promise of Automating Legal Drudgery}
\label{sec:discussion_drudgery}

One of the central promises of legal AI has long been to automate drudgery: to absorb the repetitive, low-discretion work that consumes expert labor and to free lawyers to focus on tasks that genuinely demand professional judgment \citep{ranganathan_ye_2026_ai_intensifies}. Indeed, this is the promise that legal tech vendors continue to advance today. Thomson Reuters markets its tools as helping lawyers ``make quick work of routine tasks'' \citep{thomsonreuters_legal_ai_technology}, while Harvey operates under the tagline: ``Delegate the Work. Own the Judgment.'' \citep{Harvey2026}. Even as such marketing has retreated from earlier ``hallucination-free'' claims that did not hold up under scrutiny \citep{Magesh2025, Charlotin2026a}, this shift gives the impression that while judgment-laden legal work may still be beyond the reach of automation, straightforward forms of legal drudgery can now be reliably handled by the latest AI models.

Our results call this implicit assumption into question. Insofar as Bluebooking is a useful proxy for legal drudgery---ubiquitous, heavily rule-bound, and long lamented as a rote use of highly trained time---our findings suggest that frontier LLMs have not yet fulfilled one of AI's central promises in law. Indeed, Bluebooking is, if anything, a favorable test case for the automation of drudgery. The rules are formal, public, and centralized in a single source rather than scattered across multiple overlapping regimes. The task is repetitive, low-discretion, and comparatively easy to verify. And our study gives the models a particularly charitable opportunity to succeed: in one experiment we supply the relevant \textit{Bluebook} rules directly, and in others we give allowances---such as accepting explicit markup for typeface---that are more forgiving than the ordinary conditions under which lawyers and editors actually work. If LLMs cannot reliably perform this kind of rote, structured, machine-checkable task under those conditions, broader claims that frontier models have already solved the problem of routine legal labor should be treated with caution.

Nor is it obvious that this sort of deficiency will simply disappear with more ``reasoning.'' In our experiments, reasoning mode does not produce meaningful improvement in Bluebooking performance. That is striking because reasoning-oriented models have been introduced as a major recent advance in frontier AI and have yielded substantial gains in scientific domains such as math \citep{Wang2026} and coding \citep{Yang2025}. In law, however, reasoning models seem to still fall short \citep{Zhang2026}.

Of course, our findings do not imply that AI has no value in reducing legal drudgery. In our write-on study, for example, the models perform quite well at non-Bluebooking tasks like correcting typos, and that strength materially inflates their overall scores. More promisingly, the hybrid neuro-symbolic approach we develop in Section~\ref{sec:study4} suggests that with additional engineering, LLMs can be leveraged in structured systems for rule execution and verification. But to the extent we expect AI to autonomously handle rule-bound legal tasks as a drop-in replacement for careful human attention, our results counsel significant caution.

\subsection{Bluebooking and the Limits of Law-Following AI}
\label{sec:discussing_law_following}

Our results also speak to growing calls in the AI governance community that it is necessary to develop forms of AI that are ``law-following'' in order to avoid serious AI-inflicted harms \citep{maas2025treaty, Nay2023, OKeefe2025}. Building on an earlier aspiration to make law computationally tractable by translating it into executable code \citep{buchanan1970speculation, governatori2022thirty, mccarty1977taxman, sergot1986bna}, this project rests, at least in part, on the premise that legal rules can be internalized and applied reliably by machines.

Our results suggest that this goal may be further away than those concerned about AI risk may wish. While previous research has emphasized challenges stemming from interpretive indeterminacy \citep{Shay2016, He2026}, response instability \citep{Blair-Stanek2026, Choi2027, Grimmelmann2025, Waldon2025}, and adversarial attacks \citep{Mu2024}, we consider a different problem: the sheer intricacy of many legal rules themselves. The \textit{Bluebook} is not ambiguous in the way that common-law standards or even statutory language can be, but it is extremely detailed and convoluted. If contemporary AI models are incapable of following rules of this character, then the prospects for AI systems following more contested and dynamic bodies of substantive law seem remote.

One might at first puzzle over whether our findings about AI and the \textit{Bluebook} are hard to square with other prominent findings that AI systems ace the bar exam and other legal benchmarks \citep{Guha2024, Katz2024, Martinez2026}. If models can pass what is widely regarded as the gatekeeping exam of the American legal profession, the thought runs, surely they can handle something as mechanical as citation formatting. We think the comparison cuts the other way. Those results rest primarily on multiple-choice performance, a format that probes whether a model can recognize the correct legal answer among a set of pre-specified options. Bluebooking probes something different: whether a model can produce an artifact that conforms exactly to a detailed rule set. AI models tend to struggle more on generative legal tasks than multiple-choice classification \citep{Fan2025, Martinez2024, Shi2026}, and our studies test AI's generative capacity directly. Additionally, the bar exam results have also been shown to be overstated, with GPT-4's widely cited 90th-percentile performance being closer to the median among first-time passers once properly benchmarked \citep{Martinez2024}. Thus, while models may perform well on classification benchmarks that ask them to identify correct legal outputs, it is not clear that they can robustly internalize and apply formal legal rules when asked to produce outputs from scratch.

A related puzzle concerns the growing evidence that AI assistance produces measurable productivity gains on substantive legal tasks---though gains skew toward lower performers, and may even harm those at the top \citep{Choi2024,Schwarcz2026,Chien2025}. If lawyers using LLMs are able to complete drafting, research, and analysis faster and better than those who do not, then it might seem surprising that those same models apparently cannot be trusted to follow formal legal rules on their own. We see those results as entirely consistent with ours---and as illustrating the same point. Existing productivity studies treat AI as an \textit{assistant}, a tool whose outputs are filtered by a human who retains responsibility for compliance with legal rules. Law-following AI, by contrast, requires AI to comply with legal rules \textit{autonomously}: to do, on its own, what those studies' human subjects are doing. Our Bluebooking results bear on that autonomous case, and they suggest it remains out of reach.

To be sure, our findings do not undermine the normative appeal of designing AI systems to comply with legal requirements and refuse illegal actions. But they do suggest that this approach is not yet technologically ready, contrary to claims that ``existing frontier AI systems \dots\ are already competent at reading, understanding, and reasoning about natural-language texts, including laws'' \citep{OKeefe2025}. If current systems cannot reliably follow the formal and publicly specified rules of legal citation, confidence in their ability to follow law more generally should be correspondingly tempered.

\subsection{Toward a Formal Turn in Legal AI}
\label{sec:discussion_formal_turn}

Ultimately, our findings suggest that a reorientation of legal AI development is in order. Much of the recent scholarly literature has focused on whether AI can perform discretionary, judgment-laden legal tasks: legal research, drafting, interpretation, prediction, and even adjudication. That work is important. But it has pulled attention away from a more basic question: whether AI systems can reliably perform the \textit{non-discretionary}, rule-bound work upon which legal institutions are built in the first place.

H.L.A. Hart famously described legal systems as containing a ``core of settled meaning'' governed by relatively clear rules, surrounded by a ``penumbra of debatable cases'' governed by open-textured judgments \citep{Hart1958}. Hart’s formulation, however, also came with a warning. Although he agreed that the penumbra deserves attention---``[i]ts problems are rightly the daily diet of the law schools''---he also cautioned that ``to be occupied with the penumbra is one thing, to be preoccupied with it another.'' Such preoccupation, Hart warned, is ``a source of confusion in the American legal tradition'' \citep{Hart1958}. Yet the same preoccupation now pervades legal AI.

If AI is to succeed in law, we believe that it should begin not with the penumbra but with the core. Such a reorientation would advance several of the aforementioned ambitions at once. It would better align legal AI with its ostensible promise of automating drudgery rather than judgment. It would reconnect with a foundational goal of early legal AI: to render at least some parts of law computationally tractable by translating relatively well-defined rules and procedures into code a machine can apply consistently. And it would place the more recent aspiration for law-following AI on firmer ground by beginning in domains where compliance is formal, auditable, and empirically testable.

A ``formal turn'' would not require abandoning work on interpretation, persuasion, or adjudication, but it would require changing the \textit{sequencing} of how those applications are adopted. Before deploying AI systems in discretionary, judgment-laden roles, lawyers should first establish that they can perform reliably in the non-discretionary roles that underpin those higher-order tasks. The credibility of AI in law's penumbra depends, in other words, on its reliability in law's formal core.

Our AI-augmented approach to Bluebooking illustrates what such a turn can look like in practice. Once legal rules are treated as a deterministic system to be encoded rather than a body of knowledge to be memorized and sampled from, two things change. First, AI models are no longer asked to do the part they are weakest at: faithful character-by-character rule execution under intricate exceptions. Second, the rule system becomes auditable and improvable in the way that code is auditable and improvable, rather than opaque in the way that LLM weights are opaque. The result, in our setting, is a substantial gain in compliance (\sZeroshotPooled{} to \sStructuredPooled{} pooled across models, and up to \bestStructuredAccuracy{} for \bestStructuredModel) without any change in the underlying LLMs themselves.

We do not claim that this architecture generalizes to every formal legal task; \textit{Bluebook} citation is unusually amenable to it because the rules are well specified and the inputs are easy to structure. But where those conditions are likely to hold---e.g., in tax law \citep{Jurayj2026, Yadamsuren2025}, civil legal codes \citep{Janatian2023}, administrative manuals \citep{Sunny2026}, some contract law \citep{Kant2025}, among other areas---decomposing a legal task into LLM parsing plus deterministic rule execution seems to offer a far more reliable path to automation than asking a frontier model to internalize the rules on its own.

\begin{acks}
\normalsize We thank Abdi Aidid, Jon Choi, Christoph Engel, Talia Gillis, Nicholas Mignanelli, JJ Prescott, Dave Schwartz, Justin Simard, Mirac Suzgun, and Lucia Zheng, as well as audiences at the 2025 Conference on Empirical Legal Studies and the 2025 Conference on Data Science and Law for helpful feedback. Any errors remain our own.
\end{acks}

\subsubsection*{Statements and Declarations}

The author(s) declare no potential conflicts of interest with respect to the research, authorship, and/or publication of this article. The author(s) received no financial support for the research, authorship, and/or publication of this article. The data and code for this article will be made available upon publication.

\theendnotes

\bibliographystyle{SageH}
\bibliography{citations}

\appendix

\section{Benchmark Dataset}

\subsection{Full Task Descriptions and Examples}
\label{sec:appendix_tasks}

Table~\ref{tab:task_examples} summarizes the 11 main citation tasks that appear in our benchmark dataset---with case law presented as its three citation types (basic, extended, and short form)---along with a representative example for each; Table~\ref{tab:cloze_task_examples} summarizes the 7 cloze tasks used to decompose performance on the case law task. Each task was administered in both a Bluepages variant (for practitioner documents) and a Whitepages variant (for law review articles). ``Open'' tasks require the model to generate a complete citation string from source metadata. ``Cloze'' tasks require the model to fill in a specific component of a citation (e.g., the court and date parenthetical, or a volume-reporter-page combination).

\begin{table*}[!ht]
    \centering
    \small
    \begin{tabularx}{\textwidth}{llX}
        \toprule
        \textbf{Task} & $n$ & \textbf{Example Answer} \\
        \midrule
        \multicolumn{3}{l}{\textit{Case Law}} \\
        \midrule
        Basic case citation & \fullCaseBasicQueryN{} & \textit{People v. Turner}, 342 Mich. App. 581 (2022) \\
        Extended case citation & \fullCaseExtendedQueryN{} & \textit{People v. Turner}, 342 Mich. App. 581 (2022), \textit{rev'd and remanded}, 511 Mich. 992 (2023) \\
        Case short form & \fullCaseShortFormQueryN{} & \textit{Turner}, 342 Mich. App. at 582 \\
        \midrule
        \multicolumn{3}{l}{\textit{Enacted Law}} \\
        \midrule
        Federal statutes (print) & \fedStatutesN{} & 37 U.S.C.\ \S\ 442(c) \\
        State statutes (print) & \stateStatutesN{} & Va.\ Code Ann.\ \S\ 36-51(B)--(C) (West 2019) \\
        Statutes (electronic) & \elecStatutesN{} & La.\ Child.\ Code Ann.\ art.\ 303 (Lexis+ through 2024 2d Extraordinary Sess.) \\
        Statutes short form & \statuteShortFormN{} & Nat.\ Res.\ \S\ 11.011 \\
        Administrative law & \adminLawN{} & 87 Fed. Reg. 55901 (Sept. 8, 2022) \\
        \midrule
        \multicolumn{3}{l}{\textit{Other}} \\
        \midrule
        Legislative resources & \legResourcesN{} & H.R.J.\ Res.\ 38, 115th Cong.\ (2017) \\
        Court filings & \courtFilingsN{} & Original Compl.\ \P\ 3 \\
        Secondary sources & \secondarySourcesN{} & Hayley Stillwell, \textit{Shadow Dockets Lite}, 99 Denv.\ L.\ Rev.\ 361 (2022) \\
        \bottomrule
    \end{tabularx}
    \vspace{0.3cm}
    \caption{Summary of the 11 main citation tasks in our benchmark dataset, grouped by category. Each task was evaluated in both Bluepages and Whitepages variants ($n$ reflects the combined count).}
    \label{tab:task_examples}
\end{table*}

\begin{table*}[!ht]
    \centering
    \small
    \begin{tabularx}{\textwidth}{llX}
        \toprule
        \textbf{Task} & $n$ & \textbf{Example Answer} \\
        \midrule
        Case names & \caseNamesN{} & \textit{Est.\ of Kent v.\ Kerr} \\
        Case reporters & \caseReportersN{} & \textit{Sorrentino v.\ Godinez}, 777 F.3d 410 (7th Cir.\ 2015) \\
        Case reporters (parallel) & \caseReportersParallelN{} & \textit{State v.\ Rowe}, 2022-00206 (La.\ 12/9/22), 354 So.\ 3d 1187 \\
        Case court and date & \courtDateN{} & \textit{Frost v.\ State}, 25 S.W.3d 395 (Tex.\ Crim.\ App.\ 2000) \\
        Case history & \caseHistoryN{} & \textit{Tannen v.\ Tannen}, 3 A.3d 1220 (N.J.\ Super.\ Ct.\ App.\ Div.\ 2010), \textit{aff'd}, 31 A.3d 621 (N.J.\ 2011) \\
        Case parentheticals & \caseParentheticalsN{} & \textit{Albright v.\ Oliver}, 510 U.S.\ 266, 278 (1994) (Ginsburg, J., concurring) \\
        Signals & \signalsN{} & \textit{See, e.g.}, \textit{Frost v.\ State}, 25 S.W.3d 395 (Tex.\ Crim.\ App.\ 2000) \\
        \bottomrule
    \end{tabularx}
    \vspace{0.3cm}
    \caption{Summary of the 7 cloze citation tasks in our benchmark dataset, used for decomposing performance on the main case law task. Each task was evaluated in both Bluepages and Whitepages variants ($n$ reflects the combined count).}
    \label{tab:cloze_task_examples}
\end{table*}

\subsection{Creation of the Whitepages Dataset}
\label{sec:appendix_whitepages_dataset_details}

To create the Whitepages portion of our benchmark, we scraped the online versions of recently published Harvard Law Review (HLR) articles and randomly sampled footnotes to use as our raw material. Because law review footnotes contain a different mix of authority types than the textbook exercises, we oversampled certain citation types so that the Whitepages distribution remained roughly comparable to the Bluepages benchmark, while still reflecting the greater prevalence of secondary sources in Whitepages-style academic footnotes. Each task that is covered in the Bluepages is also covered in the Whitepages, with the exception of the case reporters (parallel), which do not regularly appear in the HLR data.

From the sampled footnotes, we then extracted individual citations and mapped them onto our 18 distinct citation tasks (the 11 open-style tasks of Table~\ref{tab:task_examples} plus the 7 cloze-style tasks of Table~\ref{tab:cloze_task_examples}), excluding the parallel-reporter cloze task, for a total of 17 Whitepages task variants. Specifically, we treated the published HLR citation string as the ground-truth answer, and we generated the model query by stripping back the citation to its underlying source metadata (i.e., author/title/publication information and pincites for secondary sources; case-caption-style information of the sort displayed in Westlaw or Lexis for judicial opinions; and title/section/publisher/currentness information for statutes and regulations). We excluded purely referential short forms (e.g., \textit{id.}, \textit{supra}, \textit{infra}, and similar cross-references) and other footnote material that does not correspond to a single deterministic citation target.

\begin{figure}[!ht]
    \centering
    \includegraphics[width=\columnwidth]{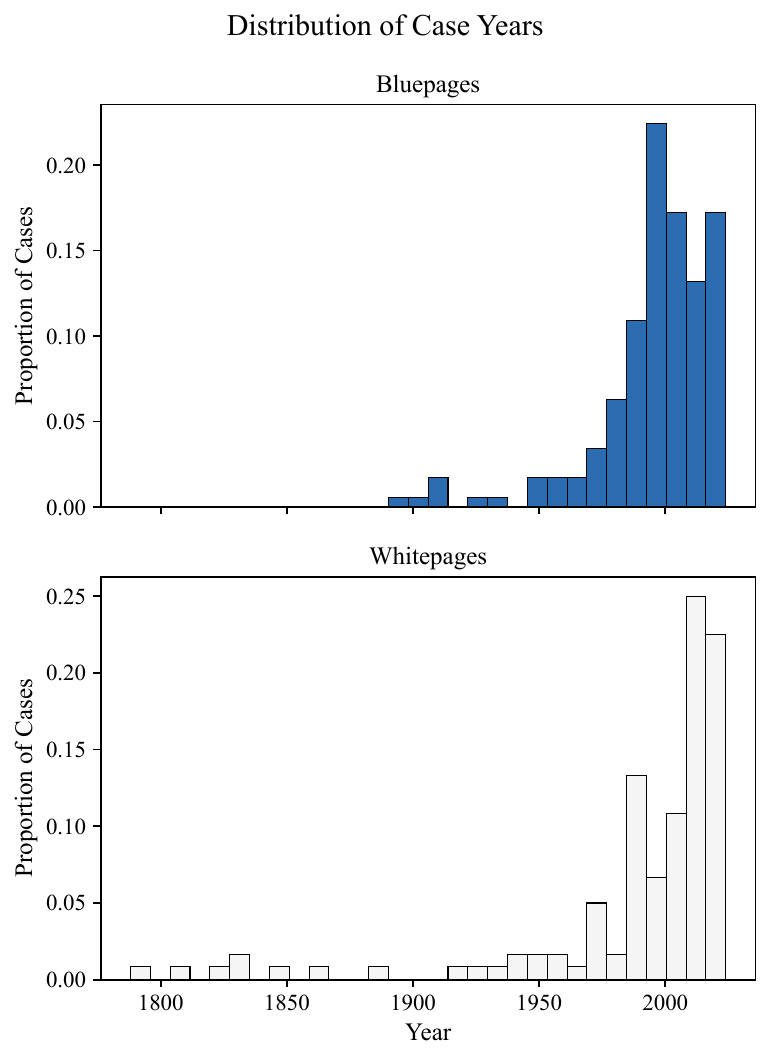}
    \caption{Temporal distribution of cited authorities in the Bluepages and Whitepages task sets.}
    \label{fig:dist_years}
\end{figure}

\begin{figure*}[!ht]
    \centering
    \includegraphics[width=\textwidth]{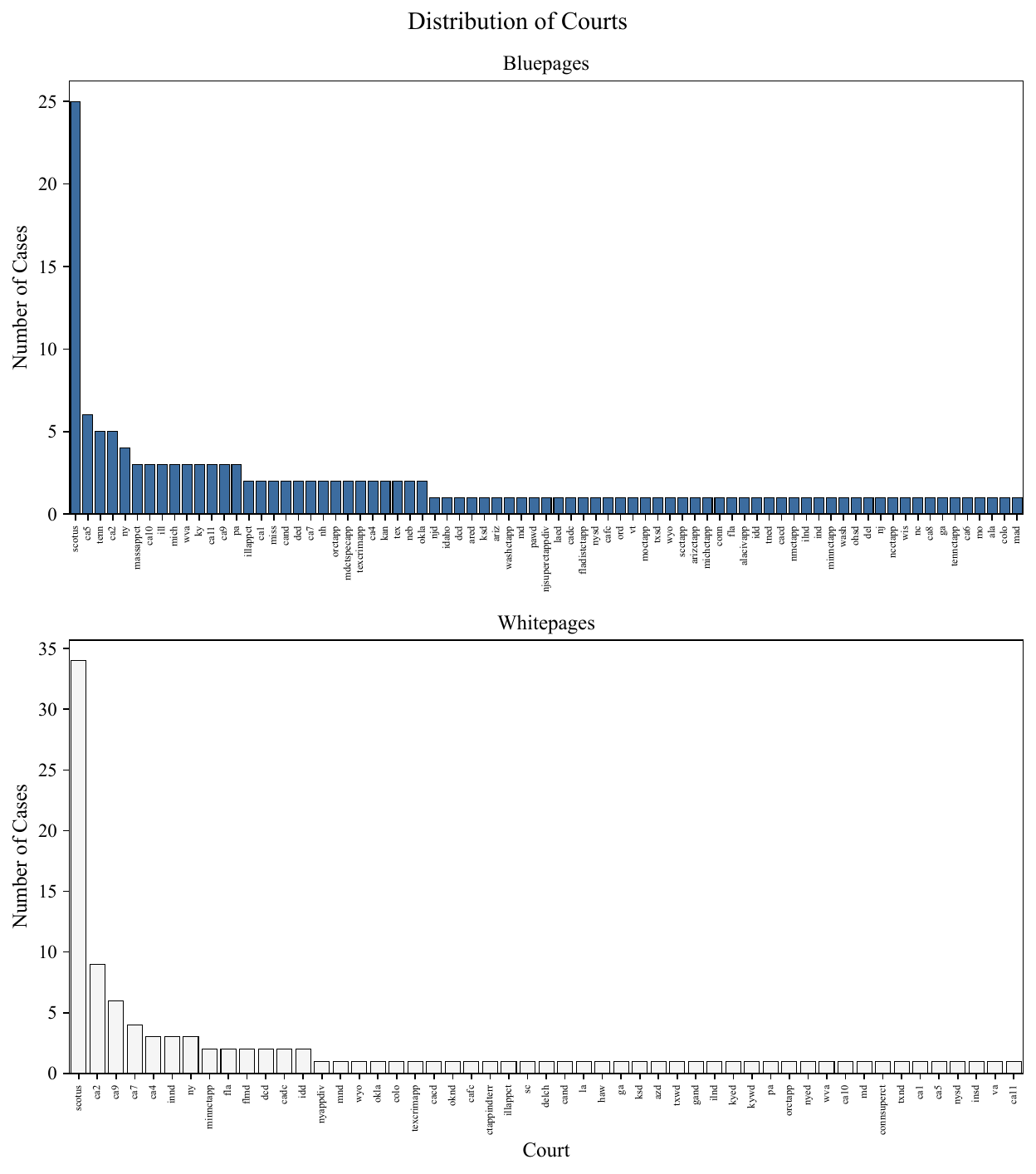}
    \vspace{0.1cm}
    \caption{Geographic distribution of cited authorities in the Bluepages and Whitepages task sets.}
    \label{fig:dist_courts}
\end{figure*}

Figures~\ref{fig:dist_years} and~\ref{fig:dist_courts} depict the temporal and geographic composition of the Bluepages and Whitepages task sets.

\section{Supplemental Benchmarking Results}

\subsection{Edit Distance by Benchmark Page}
\label{sec:appendix_edit_distance_by_page}

\begin{figure*}[!ht]
    \centering
    \begin{minipage}{0.48\textwidth}
        \centering
        \includegraphics[width=\textwidth]{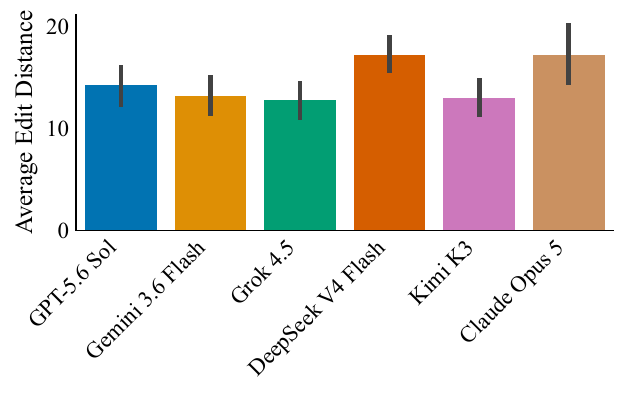}\\
        {\small (a) Bluepages}
    \end{minipage}\hfill
    \begin{minipage}{0.48\textwidth}
        \centering
        \includegraphics[width=\textwidth]{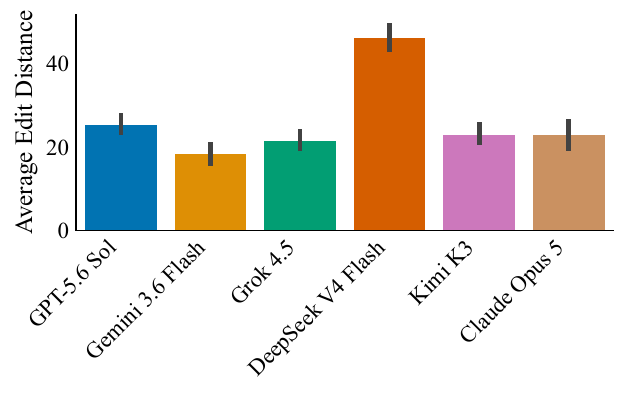}\\
        {\small (b) Whitepages}
    \end{minipage}
    \caption{Average edit distance among incorrect responses on the open formatting tasks, pooled by model and separated into the Bluepages (a) and Whitepages (b) portions of the benchmark. Compare to Figure~\ref{fig:edit_distance} in the main text. Black bars represent 95\% confidence intervals.}
    \label{fig:edit_distance_by_page}
\end{figure*}

Figure~\ref{fig:edit_distance_by_page} disaggregates the edit-distance results of Figure~\ref{fig:edit_distance} into the Bluepages and Whitepages portions of the benchmark. The pattern is qualitatively unchanged: on both pages, the typical incorrect output remains dozens of character-level edits away from the ground-truth citation.

\subsection{Typeface-Insensitive Scoring}
\label{sec:appendix_ni}

\begin{figure}[!ht]
    \centering
    \includegraphics[width=0.95\columnwidth]{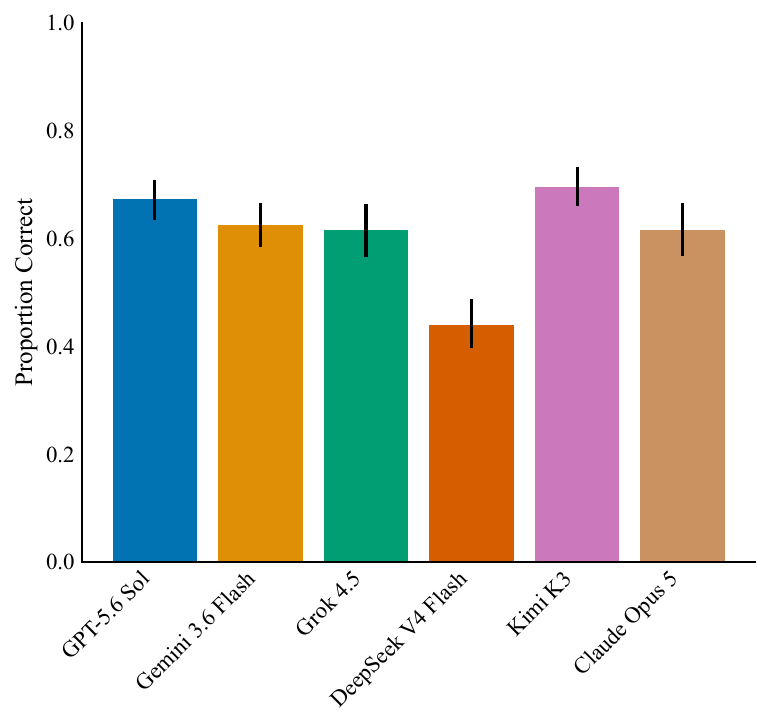}
    \caption{Accuracy by model under typeface-insensitive scoring. Compare to Figure~\ref{fig:results_by_model} (typeface-sensitive). Black bars represent 95\% confidence intervals.}
    \label{fig:results_by_model_ni}
\end{figure}

\begin{figure*}[!ht]
    \centering
    \includegraphics[width=0.9\textwidth]{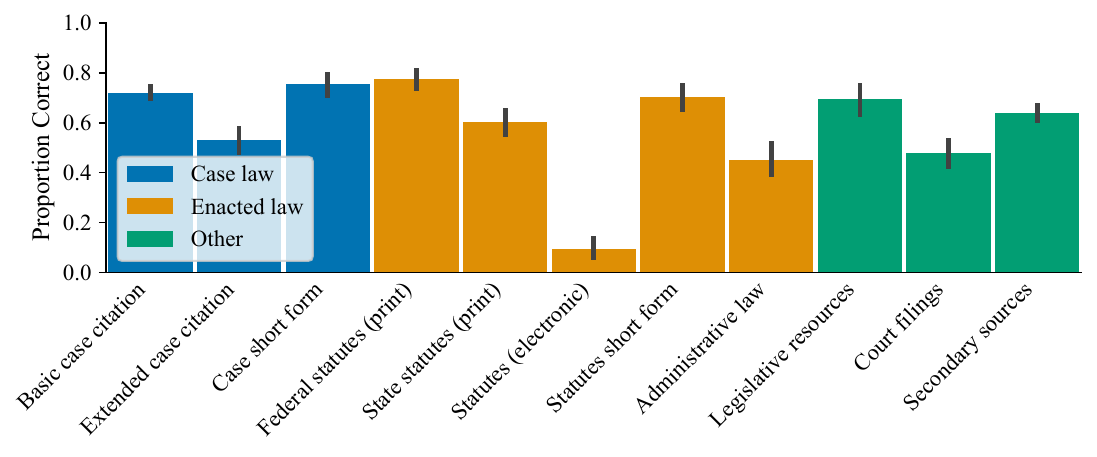}
    \caption{Accuracy by task under typeface-insensitive scoring. Compare to Figure~\ref{fig:results_by_task_case_subcats} (typeface-sensitive). Black bars represent 95\% confidence intervals.}
    \label{fig:results_by_task_ni}
\end{figure*}

All our benchmarking results in the main text enforce the \textit{Bluebook} rules regarding italicization and small caps typeface formatting. Figure~\ref{fig:results_by_model_ni} reports the accuracy for all six models under typeface-insensitive scoring instead, and Figure~\ref{fig:results_by_task_ni} shows the typeface-insensitive results pooled by model and separated by task. Full per-model, per-task typeface-insensitive results are given in Table~\ref{tab:overall_ni}, for comparison with Table~\ref{tab:overall_ti}. Ignoring typeface monotonically increases the perceived accuracy of each model, but even the typeface-insensitive results are far from perfect.

\subsection{Robustness to Answer-Key Corrections}
\label{sec:appendix_answer_key}

As described in Section~\ref{sec:study1}, our benchmark's gold answers originate from expert-produced sources: answer keys to legal-writing textbook exercises (Bluepages) and professionally Bluebooked Harvard Law Review footnotes (Whitepages). In the course of developing the deterministic formatter of Section~\ref{sec:study4}, we audited every gold answer against the governing \textit{Bluebook} rules and promoted a small number of rule-verified corrections (each documented with its rule basis) into the official answer key used throughout this paper. To confirm that our findings do not depend on these corrections, we re-scored every model's stored zero-shot responses against the original, uncorrected answer key---that is, taking the expert-produced citations entirely at face value.

Table~\ref{tab:answer_key_robustness} reports both scores for each model in our lineup ($n=$ \legacyGoldN{} responses per model---one response for each of the \nTotal{} benchmark queries). The results converge: across the \legacyGoldNModels{} models, accuracy under the original answer key differs from the official answer key by at most \legacyGoldMaxDelta{} percentage points (mean absolute difference \legacyGoldMeanDelta{} points), and no comparison or conclusion in this paper is affected by which key is used. We therefore report all results against the official (corrected) answer key, and note that the same robustness holds a fortiori for the later experiments, which reuse the same gold labels.

\begin{table}[!ht]
    \centering
    \footnotesize
    \begin{tabular}{lcc}
        \toprule
        \textbf{Model} & \textbf{Original key} & \textbf{Official key} \\
        \midrule
        \modelGpt{}      & \legacyGoldGptLegacy{}      & \legacyGoldGptOfficial{} \\
        \modelGemini{}   & \legacyGoldGeminiLegacy{}   & \legacyGoldGeminiOfficial{} \\
        \modelGrok{}     & \legacyGoldGrokLegacy{}     & \legacyGoldGrokOfficial{} \\
        \modelDeepseek{} & \legacyGoldDeepseekLegacy{} & \legacyGoldDeepseekOfficial{} \\
        \modelKimi{}     & \legacyGoldKimiLegacy{}     & \legacyGoldKimiOfficial{} \\
        \modelClaude{}   & \legacyGoldClaudeLegacy{}   & \legacyGoldClaudeOfficial{} \\
        \bottomrule
    \end{tabular}
    \vspace{0.5cm}
    \caption{Zero-shot benchmark accuracy scored against the original (uncorrected) answer key versus the official answer key incorporating rule-verified corrections ($n=$ \legacyGoldN{} responses per model; row-level accuracy over the full task set). The two keys yield converging results for every model.}
    \label{tab:answer_key_robustness}
\end{table}

\subsection{In-Context Per-Task Results}
\label{sec:appendix_in_context_bp_wp}

\begin{figure*}
    \centering
    \includegraphics[width=0.9\textwidth]{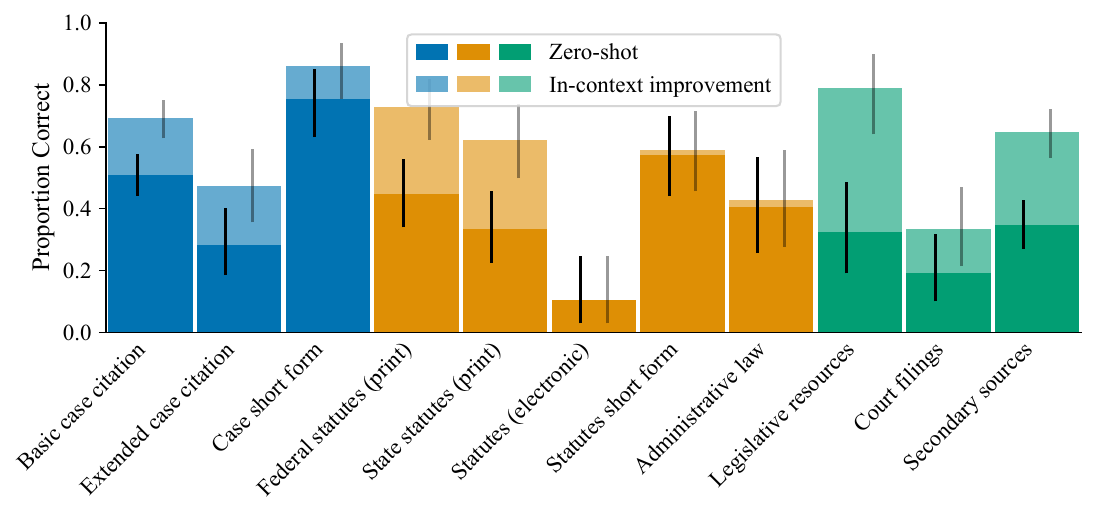}
    \vspace{0.1cm}
    \caption{Per-task results of the in-context rule-provision experiment of Section~\ref{sec:study3} (\icModelName{}, non-reasoning mode), with case law broken down into basic, extended, and short-form citations as in Figure~\ref{fig:results_by_task_case_subcats}. Dark portions of the bars represent the zero-shot performance of the model; light portions show the performance change from providing the rules in-context. Black bars represent 95\% confidence intervals.}
    \label{fig:in_context_results}
\end{figure*}

Figure~\ref{fig:in_context_results} reports the per-task results of our in-context rule provision experiment, pooling the Bluepages and Whitepages results together. Table~\ref{tab:in_context_bp_wp} further breaks out the per-task in-context learning results by page type. For reference, Table~\ref{tab:in_context_rule_list} also lists the \textit{Bluebook} rules that were provided to the model for each task.

\section{AI-Augmented Pipeline Implementation Details and Additional Results}
\label{sec:appendix_study4}

In Section~\ref{sec:study4} of the main text, we show that a hybrid neuro-symbolic system outperforms unstructured approaches to \textit{Bluebook} automation. This appendix provides additional detail on this pipeline. Section~\ref{sec:appendix_s4_pipeline} describes the LLM parsing stage, including the prompts, tool schemas, and per-provider configuration for our six frontier models. Section~\ref{sec:appendix_s4_results} reports the per-model and per-task results that our headline numbers average over, the full regression coefficient table, and several robustness checks.

\subsection{The LLM Parsing Pipeline}
\label{sec:appendix_s4_pipeline}

The parsing stage is a two-part pipeline. In step one the model is first instructed to select the citation type (from an exhaustive list of possibilities) that best matches the user's query. Then, in step two, the model is given the type-specific JSON schema for the chosen citation type and is asked to populate each field using the data drawn from the user's original natural-language query.

\subsubsection{Classifier-Stage Prompt}

The system prompt for the classifier stage is reproduced below. At runtime, the \texttt{\{citation\_type\_enum\}} placeholder is interpolated with a list of 47 different citation types, and \texttt{\{mode\}} is interpolated with either \texttt{bluepages} or \texttt{whitepages}. The prompt is short by design: the classifier's only job is type selection. The prompt template is:

\begin{quote}
\footnotesize\ttfamily
You are a Bluebook citation classifier. Your sole job is to read a user's citation question and decide which Bluebook citation TYPE it is asking about, then call the `classify' tool with that type.\\[2pt]
\# Available citation types\\[2pt]
\{citation\_type\_enum\}\\[2pt]
\# Mode\\[2pt]
The user is writing in **\{mode\}** style:\\
- `bluepages' --- practitioner / litigation document. Case names italicized; no small-caps; periodicals and books in regular type.\\
- `whitepages' --- academic / law review. Case names italicized; institutional authors and book titles in small-caps.\\[2pt]
The mode is provided for context but does NOT change classification --- it only changes how the citation will eventually be rendered.\\[2pt]
\# Rules\\[2pt]
1. Read the question. Identify what kind of source the user is asking to cite.\\
2. Pick the SINGLE best `citation\_type' from the enum above.\\
3. Call the `classify' tool with that value. Do not output any prose.
\end{quote}

The full prompt also includes detailed disambiguation guidance for short-form citations (Rules 10.9 and 12.10) and for \textit{Id.}-style follow-up cites (Rule 4.1).

\subsubsection{Parser-Stage Prompt}

The system prompt for the parser stage is a template interpolated with the chosen citation type, that type's description, the field-by-field reference, and a small set of few-shot examples to help steer the model. The template is:

\begin{quote}
\footnotesize\ttfamily
You are a Bluebook citation field extractor. The user's question contains all the information needed to assemble a Bluebook citation of type **`\{citation\_type\}`**. Your job is to extract that information into a structured object and return it via the `fill' tool.\\[2pt]
\# Mode\\[2pt]
The user is writing in **\{mode\}** style. Your output is JUST DATA --- the downstream formatter handles all typeface decisions (italics, small-caps, abbreviations) based on mode. Do not include any HTML tags or formatting in your field values.\\[2pt]
\# Citation type: `\{citation\_type\}'\\[2pt]
\{citation\_type\_description\}\\[2pt]
\# Field reference\\[2pt]
\{field\_descriptions\_block\}\\[2pt]
\# Examples\\[2pt]
\{examples\_block\}\\[2pt]
\# Rules\\[2pt]
1. Use the FULL unabbreviated form for `case\_name\_full', `reporter\_full', `court\_full'. The downstream formatter applies \textit{Bluebook} abbreviations.\\
2. Numeric fields (`volume', `page', `year', `pinpoint\_start', `pinpoint\_end') are integers when present, `null' for unpublished cases that lack them.\\
3. Pass-through what the question gives you. Do NOT invent fields the question doesn't supply.\\
4. Call the `fill' tool with your structured object. Output no prose.
\end{quote}

\subsection{Additional Results}
\label{sec:appendix_s4_results}

\subsubsection{Pipeline Ladder with Reasoning Enabled}

\begin{figure}[!ht]
    \centering
    \includegraphics[width=\columnwidth]{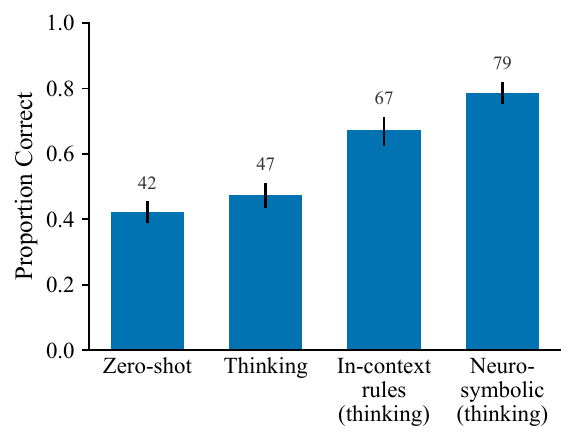}
    \caption{The pipeline comparison of Figure~\ref{fig:gpt56_ladder} with reasoning enabled for the in-context and neuro-symbolic conditions (\modelGpt{} in medium-reasoning mode; citation-frequency-weighted pooled accuracy). Black bars represent 95\% confidence intervals.}
    \label{fig:gpt56_ladder_thinking}
\end{figure}

Figure~\ref{fig:gpt56_ladder} in the main text compares the four approaches tested in this paper on \modelGpt{}, running the in-context and hybrid conditions without reasoning. Figure~\ref{fig:gpt56_ladder_thinking} reports the complementary variant in which those two conditions are instead run in medium-reasoning mode. The ordering of the four rungs is unchanged---reasoning neither rescues the in-context approach nor materially alters the hybrid pipeline's advantage.

\subsubsection{Per-Task Results}

\begin{figure*}
    \centering
    \includegraphics[width=0.9\textwidth]{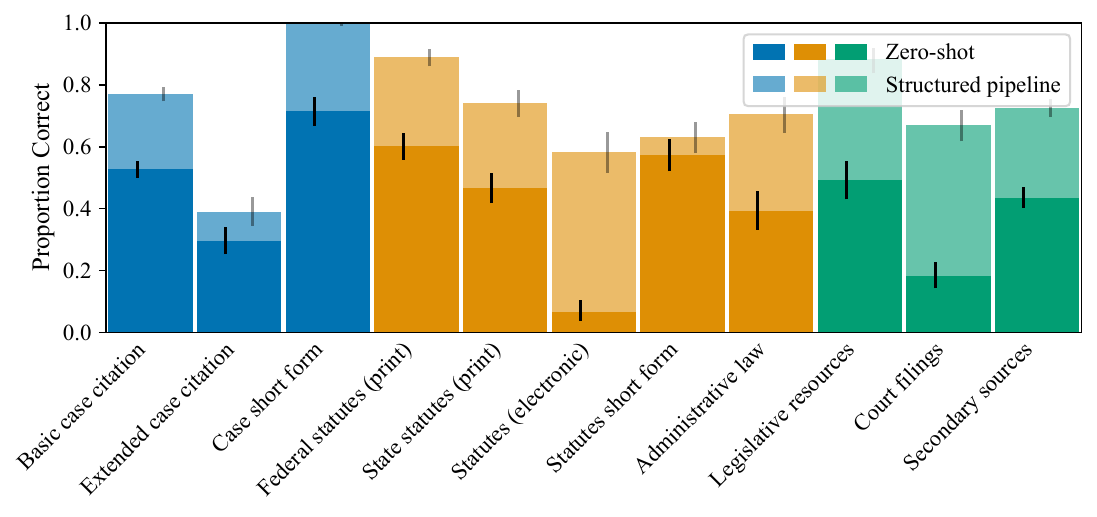}
    \vspace{0.1cm}
    \caption{The AI-augmented pipeline substantially outperforms zero-shot prompting across the open \textit{Bluebook} formatting tasks, shown with case law broken down into basic, extended, and short-form citations as in Figure~\ref{fig:results_by_task_case_subcats}. Dark portions of the bars represent the original zero-shot performance of the models; light portions show the additional accuracy gained by routing each query through the LLM parser plus deterministic formatter (all models pooled). Black bars represent 95\% confidence intervals.}
    \label{fig:structured_results_by_task}
\end{figure*}

Figure~\ref{fig:structured_results_by_task} shows the pooled by-task comparison discussed in Section~\ref{sec:study4}.

\subsubsection{Per-Model, Per-Task Accuracy}

Table~\ref{tab:s4_per_model_per_task} reports structured-pipeline accuracy for every model on each of the open citation-formatting tasks, with case law broken down into the three citation types of Figure~\ref{fig:results_by_task_case_subcats}. The pattern is consistent across the table: structured-pipeline accuracy exceeds the corresponding zero-shot baseline in nearly every cell, with the largest gains concentrated in the enacted-law and ``other'' task categories that defeated zero-shot prompting in Section~\ref{sec:study1}.

\subsubsection{Bluepages vs.\ Whitepages by Model}

The main text reports that the structured pipeline largely closes the Bluepages--Whitepages gap that we observed in Section~\ref{sec:whitepages_bluepages_gap} (\bwGap{} percentage points pooled across models). Table~\ref{tab:s4_bp_wp_by_model} shows that this gap closure occurs across all models individually and is not simply an artifact of pooling: every model's Whitepages accuracy under the structured pipeline rises substantially above its zero-shot Whitepages accuracy, and in most cases by a larger margin than its Bluepages improvement.

\begin{table*}[!ht]
    \centering
    \footnotesize
    \begin{tabular}{lcccc}
        \toprule
        \textbf{Model} & \textbf{Bluepages (zero-shot)} & \textbf{Bluepages (structured)} & \textbf{Whitepages (zero-shot)} & \textbf{Whitepages (structured)} \\
        \midrule
        \modelClaude{}    & \sClaudeZsBP{}   & \sClaudeStBP{}   & \sClaudeZsWP{}   & \sClaudeStWP{} \\
        \modelGpt{}       & \sGptZsBP{}      & \sGptStBP{}      & \sGptZsWP{}      & \sGptStWP{} \\
        \modelGemini{}    & \sGeminiZsBP{}   & \sGeminiStBP{}   & \sGeminiZsWP{}   & \sGeminiStWP{} \\
        \modelDeepseek{}  & \sDeepseekZsBP{} & \sDeepseekStBP{} & \sDeepseekZsWP{} & \sDeepseekStWP{} \\
        \modelGrok{}      & \sGrokZsBP{}     & \sGrokStBP{}     & \sGrokZsWP{}     & \sGrokStWP{} \\
        \modelKimi{}      & \sKimiZsBP{}     & \sKimiStBP{}     & \sKimiZsWP{}     & \sKimiStWP{} \\
        \bottomrule
    \end{tabular}
    \vspace{0.5cm}
    \caption{Per-model accuracy under zero-shot and structured-pipeline conditions, separately for Bluepages and Whitepages tasks.}
    \label{tab:s4_bp_wp_by_model}
\end{table*}

\subsubsection{Full Regression Table}

Table~\ref{tab:s4_regression_full} reports the full coefficient table for the logistic regression we specify in Section~\ref{sec:ai_augmented_pipeline}. The dependent variable is binary correctness; explanatory variables are condition (zero-shot vs.\ structured-pipeline), page type, their interaction, task category (\texttt{groupA} = case law, baseline; \texttt{groupB} = enacted law; \texttt{groupC} = other), and a model fixed effect with \modelClaude{} as the baseline category.

The structured-pipeline main effect ($\beta$ = \sCoefUsingStructuredOneBeta) is large and highly significant; the Whitepages-by-condition interaction ($\beta$ = \sCoefPageTypewhitepagesUsingStructuredOneBeta) is positive and significant, indicating that the structured pipeline improves Whitepages accuracy by more than it improves Bluepages accuracy---a byproduct of the formatter's deterministic typeface rendering. Model fixed effects are negative across the board (relative to \modelClaude{}, the baseline), consistent with Claude having the strongest underlying parsing accuracy in our lineup.

\begin{table}[!ht]
    \centering
    \footnotesize
    \begin{tabular}{lrrl}
        \toprule
        \textbf{Term} & \textbf{$\beta$} & \textbf{$z$} & \textbf{$p$} \\
        \midrule
        Intercept                                & \sCoefInterceptBeta            & \sCoefInterceptZ            & \sCoefInterceptP \\
        Whitepages                               & \sCoefPageTypewhitepagesBeta   & \sCoefPageTypewhitepagesZ   & \sCoefPageTypewhitepagesP \\
        Structured pipeline                      & \sCoefUsingStructuredOneBeta   & \sCoefUsingStructuredOneZ   & \sCoefUsingStructuredOneP \\
        Group B (enacted law)                    & \sCoefGroupBBeta               & \sCoefGroupBZ               & \sCoefGroupBP \\
        Group C (other)                          & \sCoefGroupCBeta               & \sCoefGroupCZ               & \sCoefGroupCP \\
        Model: DeepSeek V4 Flash                 & \sCoefModeldeepseekVFourFlashBeta      & \sCoefModeldeepseekVFourFlashZ      & \sCoefModeldeepseekVFourFlashP \\
        Model: Gemini 3.6 Flash                  & \sCoefModelgeminiThreeSixFlashBeta     & \sCoefModelgeminiThreeSixFlashZ     & \sCoefModelgeminiThreeSixFlashP \\
        Model: \modelGpt{}                       & \sCoefModelgptFiveSixSolThinkNoneBeta  & \sCoefModelgptFiveSixSolThinkNoneZ  & \sCoefModelgptFiveSixSolThinkNoneP \\
        Model: Grok 4.5                          & \sCoefModelgrokFourFiveBeta            & \sCoefModelgrokFourFiveZ            & \sCoefModelgrokFourFiveP \\
        Model: Kimi K3                           & \sCoefModelkimiKThreeBeta              & \sCoefModelkimiKThreeZ              & \sCoefModelkimiKThreeP \\
        Whitepages $\times$ Structured pipeline  & \sCoefPageTypewhitepagesUsingStructuredOneBeta & \sCoefPageTypewhitepagesUsingStructuredOneZ & \sCoefPageTypewhitepagesUsingStructuredOneP \\
        \bottomrule
    \end{tabular}
    \vspace{0.5cm}
    \caption{Full coefficient table for the logistic regression reported in Section~\ref{sec:study4} testing the AI-augmented pipeline. Baseline categories: Group A (case law), Bluepages, zero-shot, \modelClaude{}. Standard errors are clustered by query (\sNClusters{} clusters); $n$ = \sN{} observations total.}
    \label{tab:s4_regression_full}
\end{table}


\begin{table*}[!ht]
    \centering
    \setlength{\tabcolsep}{4pt}
    \begin{tabular}{lcccccc}
        \toprule
        \textbf{Task} & \textbf{Claude} & \textbf{\modelGpt} & \textbf{Gemini} & \textbf{DeepSeek} & \textbf{Grok} & \textbf{Kimi} \\
        \midrule
        Basic case citation       & \sClaudeFullCaseBasicSt{}          & \sGptFullCaseBasicSt{}          & \sGeminiFullCaseBasicSt{}          & \sDeepseekFullCaseBasicSt{}          & \sGrokFullCaseBasicSt{}          & \sKimiFullCaseBasicSt{} \\
        Extended case citation    & \sClaudeFullCaseExtendedSt{}       & \sGptFullCaseExtendedSt{}       & \sGeminiFullCaseExtendedSt{}       & \sDeepseekFullCaseExtendedSt{}       & \sGrokFullCaseExtendedSt{}       & \sKimiFullCaseExtendedSt{} \\
        Case short form           & \sClaudeFullCaseShortFormSt{}      & \sGptFullCaseShortFormSt{}      & \sGeminiFullCaseShortFormSt{}      & \sDeepseekFullCaseShortFormSt{}      & \sGrokFullCaseShortFormSt{}      & \sKimiFullCaseShortFormSt{} \\
        \midrule
        Federal statutes (print)  & \sClaudeFedStatutesSt{}            & \sGptFedStatutesSt{}            & \sGeminiFedStatutesSt{}            & \sDeepseekFedStatutesSt{}            & \sGrokFedStatutesSt{}            & \sKimiFedStatutesSt{} \\
        State statutes (print)    & \sClaudeStateStatutesSt{}          & \sGptStateStatutesSt{}          & \sGeminiStateStatutesSt{}          & \sDeepseekStateStatutesSt{}          & \sGrokStateStatutesSt{}          & \sKimiStateStatutesSt{} \\
        Statutes (electronic)     & \sClaudeElecStatutesSt{}           & \sGptElecStatutesSt{}           & \sGeminiElecStatutesSt{}           & \sDeepseekElecStatutesSt{}           & \sGrokElecStatutesSt{}           & \sKimiElecStatutesSt{} \\
        Statutes short form       & \sClaudeStatuteShortFormSt{}       & \sGptStatuteShortFormSt{}       & \sGeminiStatuteShortFormSt{}       & \sDeepseekStatuteShortFormSt{}       & \sGrokStatuteShortFormSt{}       & \sKimiStatuteShortFormSt{} \\
        Administrative law        & \sClaudeAdminLawSt{}               & \sGptAdminLawSt{}               & \sGeminiAdminLawSt{}               & \sDeepseekAdminLawSt{}               & \sGrokAdminLawSt{}               & \sKimiAdminLawSt{} \\
        \midrule
        Legislative resources     & \sClaudeLegResourcesSt{}           & \sGptLegResourcesSt{}           & \sGeminiLegResourcesSt{}           & \sDeepseekLegResourcesSt{}           & \sGrokLegResourcesSt{}           & \sKimiLegResourcesSt{} \\
        Court filings             & \sClaudeCourtFilingsSt{}           & \sGptCourtFilingsSt{}           & \sGeminiCourtFilingsSt{}           & \sDeepseekCourtFilingsSt{}           & \sGrokCourtFilingsSt{}           & \sKimiCourtFilingsSt{} \\
        Secondary sources         & \sClaudeSecondarySourcesSt{}       & \sGptSecondarySourcesSt{}       & \sGeminiSecondarySourcesSt{}       & \sDeepseekSecondarySourcesSt{}       & \sGrokSecondarySourcesSt{}       & \sKimiSecondarySourcesSt{} \\
        \bottomrule
    \end{tabular}
    \vspace{0.5cm}
    \caption{Per-model, per-task structured-pipeline accuracy. Per-model zero-shot baselines are reported in Table~\ref{tab:s4_bp_wp_by_model}; pooled per-task zero-shot baselines are shown in Figure~\ref{fig:results_by_task_case_subcats}.}
    \label{tab:s4_per_model_per_task}
\end{table*}

\begin{table*}[!ht]
    \centering
    \begin{tabular}{lcccc}
        \toprule
        & \multicolumn{2}{c}{\textbf{Bluepages}} & \multicolumn{2}{c}{\textbf{Whitepages}} \\
        \cmidrule(lr){2-3}\cmidrule(lr){4-5}
        \textbf{Task} & Zero-shot & In-context & Zero-shot & In-context \\
        \midrule
        Basic case citation & \icFullcasebasicZsBP & \icFullcasebasicIcBP & \icFullcasebasicZsWP & \icFullcasebasicIcWP \\
        Extended case citation & \icFullcaseextendedZsBP & \icFullcaseextendedIcBP & \icFullcaseextendedZsWP & \icFullcaseextendedIcWP \\
        Case short form & \icFullcaseshortformZsBP & \icFullcaseshortformIcBP & \icFullcaseshortformZsWP & \icFullcaseshortformIcWP \\
        \midrule
        Federal statutes (print) & \icFedstatutesZsBP & \icFedstatutesIcBP & \icFedstatutesZsWP & \icFedstatutesIcWP \\
        State statutes (print) & \icStatestatutesZsBP & \icStatestatutesIcBP & \icStatestatutesZsWP & \icStatestatutesIcWP \\
        Statutes (electronic) & \icElecstatutesZsBP & \icElecstatutesIcBP & \icElecstatutesZsWP & \icElecstatutesIcWP \\
        Statutes short form & \icStatuteshortformZsBP & \icStatuteshortformIcBP & \icStatuteshortformZsWP & \icStatuteshortformIcWP \\
        Administrative law & \icAdminlawZsBP & \icAdminlawIcBP & \icAdminlawZsWP & \icAdminlawIcWP \\
        \midrule
        Legislative resources & \icLegresourcesZsBP & \icLegresourcesIcBP & \icLegresourcesZsWP & \icLegresourcesIcWP \\
        Court filings & \icCourtfilingsZsBP & \icCourtfilingsIcBP & \icCourtfilingsZsWP & \icCourtfilingsIcWP \\
        Secondary sources & \icSecondarysourcesZsBP & \icSecondarysourcesIcBP & \icSecondarysourcesZsWP & \icSecondarysourcesIcWP \\
        \bottomrule
    \end{tabular}
    \vspace{0.3cm}
    \caption{Per-task accuracy for \icModelName{} under zero-shot and rule-supplied (in-context) prompting, split by page type. The pooled view---combining both page types---appears in Figure~\ref{fig:in_context_results}.}
    \label{tab:in_context_bp_wp}
\end{table*}

\begin{table*}[ht]
\centering
\begin{tabular}{>{\raggedright}p{4cm} >{\raggedright\arraybackslash}p{12cm}}
\toprule
\textbf{Task} & \textbf{Rules} \\
\midrule
Basic case citation & 1.5, 4.1, 4.2, 10.2, 10.3, 10.4, 10.5, 10.6, 10.7, 10.8, 10.9, B1.3, B4, B10.1, B10.2, T1.3, T6, T8 \\
Extended case citation & 1.5, 4.1, 4.2, 10.2, 10.3, 10.4, 10.5, 10.6, 10.7, 10.8, 10.9, B1.3, B4, B10.1, B10.2, T1.3, T6, T8 \\
Short form case citation & 1.5, 4.1, 4.2, 10.2, 10.3, 10.4, 10.5, 10.6, 10.7, 10.8, 10.9, B1.3, B4, B10.1, B10.2, T1.3, T6, T8 \\
\midrule
Federal statutes (print) & 11, 12, 12.1, 12.2, 12.3, 12.4, 12.6, 12.7, 12.8, 12.9, B3, B11, B12.1, T16, T1.1 \\
State statutes (print)   & 11, 12, 12.1, 12.2, 12.3, 12.4, 12.6, 12.7, 12.8, 12.9, B3, B11, B12.1, T16, T1.3 \\
Statutes (electronic)    & 12.3, 12.5, 18.2, T16 \\
Statutes short form      & 4.1, 4.2, 6.2, 11, 12.1, 12.2, 12.3, 12.10, B4, B11, B12.2, T16 \\
Administrative law       & 12.9, 14, B12.1, B12.2, B14, T1.2 \\
\midrule
Legislative resources & 12.4, 13, B12.1, B12.2, B13, B17.1, B17.2 \\
Court filings         & 10.8, B17.1, B17.2, BT1 \\
Secondary sources     & 15, 16, 18, B15.1, B16.1, B18.1 \\
Signals               & 1.2, 1.3, 1.4, B1.2 \\
\bottomrule
\end{tabular}
\vspace{0.3cm}
\caption{\textit{Bluebook} rules provided to the model in the in-context learning experiment (Section~\ref{sec:study3}). In addition to the task-specific rules listed in the table, the model was also always given the general rules 2, 3, 6, 7, 8, B2, B3, B6, B7, and B8.}
\label{tab:in_context_rule_list}
\end{table*}

\begin{table*}[!ht]
    \centering
    \small
    \input{tables/per_task_overall_typeface_sensitive.tex}
    \vspace{0.3cm}
    \caption{Per-task accuracy under typeface-sensitive scoring, the primary metric reported in the main text. The ``Pooled (weighted)'' row is each model's headline pooled accuracy: the citation-frequency-weighted statistic over the open formatting tasks (cloze rows excluded). Compare to Table~\ref{tab:overall_ni} (typeface-insensitive).}
    \label{tab:overall_ti}
\end{table*}

\begin{table*}[!ht]
    \centering
    \small
    \input{tables/per_task_overall_typeface_insensitive.tex}
    \vspace{0.3cm}
    \caption{Per-task accuracy under typeface-insensitive scoring. The ``Pooled (weighted)'' row is the citation-frequency-weighted pooled statistic over the open formatting tasks (cloze rows excluded). Compare to Table~\ref{tab:overall_ti} (typeface-sensitive).}
    \label{tab:overall_ni}
\end{table*}

\end{document}

%% file: generated_values.tex
\newcommand{\nTotal}{2,058}
\newcommand{\nPooled}{914}
\newcommand{\nClozeTotal}{1,144}
\newcommand{\pooledAccuracy}{42.6\%}

\newcommand{\hlrWeightCases}{33.1\%}
\newcommand{\hlrWeightStatutes}{10.0\%}
\newcommand{\hlrWeightOther}{56.9\%}
\newcommand{\hlrN}{5,322}
\newcommand{\hlrCoverageRate}{98.3\%}
\newcommand{\hlrUncoveredN}{90}
\newcommand{\bestModelName}{Claude Opus 5}
\newcommand{\bestModelAccuracy}{51.3\%}
\newcommand{\worstModelName}{DeepSeek V4 Flash}
\newcommand{\worstModelAccuracy}{22.7\%}

\newcommand{\claudeAccuracy}{51.3\%}

\newcommand{\modelGpt}{GPT-5.6 Sol}
\newcommand{\modelGemini}{Gemini 3.6 Flash}
\newcommand{\modelGrok}{Grok 4.5}
\newcommand{\modelDeepseek}{DeepSeek V4 Flash}
\newcommand{\modelKimi}{Kimi K3}
\newcommand{\modelClaude}{Claude Opus 5}

\newcommand{\gptAccuracyBP}{66.3\%}

\newcommand{\geminiAccuracyBP}{63.6\%}

\newcommand{\grokAccuracyBP}{62.9\%}

\newcommand{\deepseekAccuracyBP}{29.5\%}

\newcommand{\kimiAccuracyBP}{63.2\%}

\newcommand{\claudeAccuracyBP}{60.5\%}

\newcommand{\gptAccuracyWP}{18.1\%}

\newcommand{\geminiAccuracyWP}{34.8\%}

\newcommand{\grokAccuracyWP}{30.0\%}

\newcommand{\deepseekAccuracyWP}{16.0\%}

\newcommand{\kimiAccuracyWP}{24.0\%}

\newcommand{\claudeAccuracyWP}{42.1\%}

\newcommand{\bluepagesAccuracy}{57.7\%}

\newcommand{\whitepagesAccuracy}{27.5\%}

\newcommand{\bwGap}{30.2}
\newcommand{\nWhitepagesPerModel}{1,007}
\newcommand{\nBluepagesPerModel}{1,051}

\newcommand{\fedStatutesAccuracy}{60.2\%}

\newcommand{\fedStatutesN}{85}
\newcommand{\stateStatutesAccuracy}{46.6\%}

\newcommand{\stateStatutesN}{69}
\newcommand{\elecStatutesAccuracy}{6.6\%}

\newcommand{\elecStatutesN}{38}
\newcommand{\statuteShortFormAccuracy}{57.4\%}

\newcommand{\statuteShortFormN}{61}
\newcommand{\adminLawAccuracy}{39.3\%}

\newcommand{\adminLawN}{42}
\newcommand{\legResourcesAccuracy}{49.2\%}

\newcommand{\legResourcesN}{43}
\newcommand{\courtFilingsAccuracy}{18.1\%}

\newcommand{\courtFilingsN}{57}
\newcommand{\secondarySourcesAccuracy}{43.6\%}

\newcommand{\secondarySourcesN}{150}

\newcommand{\caseNamesN}{331}
\newcommand{\caseReportersAccuracy}{57.7\%}

\newcommand{\caseReportersN}{285}

\newcommand{\caseReportersParallelN}{25}
\newcommand{\courtDateAccuracy}{58.0\%}

\newcommand{\courtDateN}{305}
\newcommand{\caseHistoryAccuracy}{40.4\%}

\newcommand{\caseHistoryN}{40}
\newcommand{\caseParentheticalsAccuracy}{42.0\%}

\newcommand{\caseParentheticalsN}{50}
\newcommand{\signalsAccuracy}{36.4\%}

\newcommand{\signalsN}{108}

\newcommand{\fullCaseBasicAccuracy}{52.7\%}

\newcommand{\fullCaseExtendedAccuracy}{29.5\%}

\newcommand{\fullCaseShortFormAccuracy}{71.5\%}

\newcommand{\fullCaseBasicQueryN}{230}
\newcommand{\fullCaseExtendedQueryN}{74}
\newcommand{\fullCaseShortFormQueryN}{65}

\newcommand{\thinkingModelName}{GPT-5.6 Sol}
\newcommand{\gptNonthinkingAccuracy}{42.2\%}

\newcommand{\gptNonthinkingAccuracyBP}{66.3\%}

\newcommand{\gptNonthinkingAccuracyWP}{18.1\%}

\newcommand{\gptThinkingAccuracy}{47.5\%}

\newcommand{\gptThinkingAccuracyBP}{69.7\%}

\newcommand{\gptThinkingAccuracyWP}{25.3\%}

\newcommand{\gptThinkingAccuracyWPNI}{64.5\%}

\newcommand{\claudeEditDist}{21.1}

\newcommand{\pooledEditDistNI}{17.4}

\newcommand{\icModelName}{GPT-5.6 Sol}
\newcommand{\icZeroshotAccuracy}{42.2\%}

\newcommand{\icIncontextAccuracy}{63.1\%}

\newcommand{\icGain}{20.9}
\newcommand{\icZeroshotAccuracyBP}{66.3\%}

\newcommand{\icIncontextAccuracyBP}{79.2\%}

\newcommand{\icZeroshotAccuracyWP}{18.1\%}

\newcommand{\icIncontextAccuracyWP}{47.1\%}

\newcommand{\icFullcasebasicZsBP}{76.0\%}
\newcommand{\icFullcasebasicIcBP}{80.5\%}
\newcommand{\icFullcaseextendedZsBP}{67.7\%}
\newcommand{\icFullcaseextendedIcBP}{83.9\%}
\newcommand{\icFullcaseshortformZsBP}{71.0\%}
\newcommand{\icFullcaseshortformIcBP}{93.5\%}
\newcommand{\icFedstatutesZsBP}{56.8\%}
\newcommand{\icFedstatutesIcBP}{97.3\%}
\newcommand{\icStatestatutesZsBP}{58.8\%}
\newcommand{\icStatestatutesIcBP}{76.5\%}
\newcommand{\icElecstatutesZsBP}{15.0\%}
\newcommand{\icElecstatutesIcBP}{10.0\%}
\newcommand{\icStatuteshortformZsBP}{53.6\%}
\newcommand{\icStatuteshortformIcBP}{57.1\%}
\newcommand{\icAdminlawZsBP}{46.7\%}
\newcommand{\icAdminlawIcBP}{53.3\%}
\newcommand{\icLegresourcesZsBP}{100.0\%}
\newcommand{\icLegresourcesIcBP}{87.5\%}
\newcommand{\icCourtfilingsZsBP}{45.0\%}
\newcommand{\icCourtfilingsIcBP}{85.0\%}
\newcommand{\icSecondarysourcesZsBP}{51.0\%}
\newcommand{\icSecondarysourcesIcBP}{69.4\%}

\newcommand{\icFullcasebasicZsWP}{0.0\%}
\newcommand{\icFullcasebasicIcWP}{46.1\%}
\newcommand{\icFullcaseextendedZsWP}{0.0\%}
\newcommand{\icFullcaseextendedIcWP}{20.9\%}
\newcommand{\icFullcaseshortformZsWP}{79.4\%}
\newcommand{\icFullcaseshortformIcWP}{79.4\%}
\newcommand{\icFedstatutesZsWP}{35.4\%}
\newcommand{\icFedstatutesIcWP}{54.2\%}
\newcommand{\icStatestatutesZsWP}{8.6\%}
\newcommand{\icStatestatutesIcWP}{48.6\%}
\newcommand{\icElecstatutesZsWP}{5.6\%}
\newcommand{\icElecstatutesIcWP}{11.1\%}
\newcommand{\icStatuteshortformZsWP}{60.6\%}
\newcommand{\icStatuteshortformIcWP}{60.6\%}
\newcommand{\icAdminlawZsWP}{37.0\%}
\newcommand{\icAdminlawIcWP}{37.0\%}
\newcommand{\icLegresourcesZsWP}{17.1\%}
\newcommand{\icLegresourcesIcWP}{77.1\%}
\newcommand{\icCourtfilingsZsWP}{5.4\%}
\newcommand{\icCourtfilingsIcWP}{5.4\%}
\newcommand{\icSecondarysourcesZsWP}{26.7\%}
\newcommand{\icSecondarysourcesIcWP}{62.4\%}

\newcommand{\icFullcasebasicZs}{50.9\%}
\newcommand{\icFullcasebasicIc}{69.1\%}

\newcommand{\icFedstatutesZs}{44.7\%}
\newcommand{\icFedstatutesIc}{72.9\%}
\newcommand{\icStatestatutesZs}{33.3\%}
\newcommand{\icStatestatutesIc}{62.3\%}

\newcommand{\icElecstatutesIc}{10.5\%}
\newcommand{\icStatuteshortformZs}{57.4\%}
\newcommand{\icStatuteshortformIc}{59.0\%}
\newcommand{\icAdminlawZs}{40.5\%}
\newcommand{\icAdminlawIc}{42.9\%}
\newcommand{\icLegresourcesZs}{32.6\%}
\newcommand{\icLegresourcesIc}{79.1\%}

\newcommand{\icSecondarysourcesZs}{34.7\%}
\newcommand{\icSecondarysourcesIc}{64.7\%}

\newcommand{\bbP}{$p < .001$}

\newcommand{\bbInterP}{$p < .001$}
\newcommand{\bbBetaNI}{0.62}

\newcommand{\bbPNI}{$p < .001$}
\newcommand{\bbInterBetaNI}{-0.47}

\newcommand{\bbInterPNI}{$p < .001$}

\newcommand{\formatterCeilingAccuracy}{100.00\%}

\newcommand{\formatterCeilingBP}{100.00\%}
\newcommand{\formatterCeilingWP}{100.00\%}

\newcommand{\formatterHlrAccuracy}{98.2\%}
\newcommand{\formatterHlrN}{228}

\newcommand{\sGptZsBP}{66.3\%}
\newcommand{\sGptZsWP}{18.1\%}

\newcommand{\sGptStBP}{84.0\%}
\newcommand{\sGptStWP}{73.8\%}

\newcommand{\sGeminiZsBP}{63.6\%}
\newcommand{\sGeminiZsWP}{34.8\%}

\newcommand{\sGeminiStBP}{82.0\%}
\newcommand{\sGeminiStWP}{67.7\%}

\newcommand{\sGrokZsBP}{62.9\%}
\newcommand{\sGrokZsWP}{30.0\%}

\newcommand{\sGrokStBP}{86.8\%}
\newcommand{\sGrokStWP}{74.8\%}

\newcommand{\sDeepseekZs}{22.7\%}
\newcommand{\sDeepseekZsBP}{29.5\%}
\newcommand{\sDeepseekZsWP}{16.0\%}
\newcommand{\sDeepseekSt}{77.0\%}
\newcommand{\sDeepseekStBP}{86.4\%}
\newcommand{\sDeepseekStWP}{67.7\%}

\newcommand{\sKimiZsBP}{63.2\%}
\newcommand{\sKimiZsWP}{24.0\%}

\newcommand{\sKimiStBP}{60.4\%}
\newcommand{\sKimiStWP}{45.7\%}

\newcommand{\sClaudeZs}{51.3\%}
\newcommand{\sClaudeZsBP}{60.5\%}
\newcommand{\sClaudeZsWP}{42.1\%}

\newcommand{\sClaudeStBP}{91.4\%}
\newcommand{\sClaudeStWP}{79.7\%}

\newcommand{\sZeroshotPooled}{42.6\%}

\newcommand{\sZeroshotPooledBP}{57.7\%}
\newcommand{\sZeroshotPooledWP}{27.5\%}
\newcommand{\sStructuredPooled}{75.0\%}

\newcommand{\sStructuredPooledBP}{81.8\%}
\newcommand{\sStructuredPooledWP}{68.2\%}
\newcommand{\sGain}{32.4}
\newcommand{\sGainBP}{24.2}
\newcommand{\sGainWP}{40.7}
\newcommand{\bestStructuredAccuracy}{85.5\%}
\newcommand{\bestStructuredModel}{Claude Opus 5}

\newcommand{\sFullCaseBasicZs}{52.7\%}
\newcommand{\sFullCaseBasicSt}{77.0\%}
\newcommand{\sFullCaseExtendedZs}{29.5\%}
\newcommand{\sFullCaseExtendedSt}{39.0\%}

\newcommand{\sStateStatutesZs}{46.6\%}
\newcommand{\sStateStatutesSt}{74.2\%}

\newcommand{\sElecStatutesSt}{58.3\%}

\newcommand{\sLegResourcesZs}{49.2\%}
\newcommand{\sLegResourcesSt}{88.4\%}
\newcommand{\sCourtFilingsZs}{18.1\%}
\newcommand{\sCourtFilingsSt}{67.0\%}
\newcommand{\sSecondarySourcesZs}{43.6\%}
\newcommand{\sSecondarySourcesSt}{72.7\%}

\newcommand{\sN}{10,968}
\newcommand{\sNClusters}{914}

\newcommand{\sP}{$p < .001$}

\newcommand{\sClaudeFullCaseBasicSt}{87.4\%}

\newcommand{\sClaudeFullCaseExtendedSt}{67.6\%}

\newcommand{\sClaudeFullCaseShortFormSt}{100.0\%}

\newcommand{\sClaudeFedStatutesSt}{97.6\%}

\newcommand{\sClaudeStateStatutesSt}{85.5\%}

\newcommand{\sClaudeElecStatutesSt}{71.1\%}

\newcommand{\sClaudeStatuteShortFormSt}{83.6\%}

\newcommand{\sClaudeAdminLawSt}{81.0\%}

\newcommand{\sClaudeLegResourcesSt}{95.3\%}

\newcommand{\sClaudeCourtFilingsSt}{77.2\%}

\newcommand{\sClaudeSecondarySourcesSt}{84.0\%}

\newcommand{\sGptFullCaseBasicSt}{81.7\%}

\newcommand{\sGptFullCaseExtendedSt}{45.9\%}

\newcommand{\sGptFullCaseShortFormSt}{100.0\%}

\newcommand{\sGptFedStatutesSt}{94.1\%}

\newcommand{\sGptStateStatutesSt}{72.5\%}

\newcommand{\sGptElecStatutesSt}{57.9\%}

\newcommand{\sGptStatuteShortFormSt}{75.4\%}

\newcommand{\sGptAdminLawSt}{73.8\%}

\newcommand{\sGptLegResourcesSt}{95.3\%}

\newcommand{\sGptCourtFilingsSt}{70.2\%}

\newcommand{\sGptSecondarySourcesSt}{79.3\%}

\newcommand{\sGeminiFullCaseBasicSt}{73.5\%}

\newcommand{\sGeminiFullCaseExtendedSt}{21.6\%}

\newcommand{\sGeminiFullCaseShortFormSt}{100.0\%}

\newcommand{\sGeminiFedStatutesSt}{85.9\%}

\newcommand{\sGeminiStateStatutesSt}{72.5\%}

\newcommand{\sGeminiElecStatutesSt}{55.3\%}

\newcommand{\sGeminiStatuteShortFormSt}{60.7\%}

\newcommand{\sGeminiAdminLawSt}{69.0\%}

\newcommand{\sGeminiLegResourcesSt}{90.7\%}

\newcommand{\sGeminiCourtFilingsSt}{70.2\%}

\newcommand{\sGeminiSecondarySourcesSt}{77.3\%}

\newcommand{\sDeepseekFullCaseBasicSt}{76.5\%}

\newcommand{\sDeepseekFullCaseExtendedSt}{47.3\%}

\newcommand{\sDeepseekFullCaseShortFormSt}{100.0\%}

\newcommand{\sDeepseekFedStatutesSt}{95.3\%}

\newcommand{\sDeepseekStateStatutesSt}{81.2\%}

\newcommand{\sDeepseekElecStatutesSt}{73.7\%}

\newcommand{\sDeepseekStatuteShortFormSt}{73.8\%}

\newcommand{\sDeepseekAdminLawSt}{78.6\%}

\newcommand{\sDeepseekLegResourcesSt}{83.7\%}

\newcommand{\sDeepseekCourtFilingsSt}{70.2\%}

\newcommand{\sDeepseekSecondarySourcesSt}{72.7\%}

\newcommand{\sGrokFullCaseBasicSt}{83.0\%}

\newcommand{\sGrokFullCaseExtendedSt}{51.4\%}

\newcommand{\sGrokFullCaseShortFormSt}{100.0\%}

\newcommand{\sGrokFedStatutesSt}{97.6\%}

\newcommand{\sGrokStateStatutesSt}{84.1\%}

\newcommand{\sGrokElecStatutesSt}{60.5\%}

\newcommand{\sGrokStatuteShortFormSt}{77.0\%}

\newcommand{\sGrokAdminLawSt}{78.6\%}

\newcommand{\sGrokLegResourcesSt}{93.0\%}

\newcommand{\sGrokCourtFilingsSt}{75.4\%}

\newcommand{\sGrokSecondarySourcesSt}{76.7\%}

\newcommand{\sKimiFullCaseBasicSt}{59.6\%}

\newcommand{\sKimiFullCaseExtendedSt}{0.0\%}

\newcommand{\sKimiFullCaseShortFormSt}{100.0\%}

\newcommand{\sKimiFedStatutesSt}{63.5\%}

\newcommand{\sKimiStateStatutesSt}{49.3\%}

\newcommand{\sKimiElecStatutesSt}{31.6\%}

\newcommand{\sKimiStatuteShortFormSt}{8.2\%}

\newcommand{\sKimiAdminLawSt}{42.9\%}

\newcommand{\sKimiLegResourcesSt}{72.1\%}

\newcommand{\sKimiCourtFilingsSt}{38.6\%}

\newcommand{\sKimiSecondarySourcesSt}{46.0\%}

\newcommand{\sCoefInterceptBeta}{1.03}
\newcommand{\sCoefInterceptZ}{10.77}
\newcommand{\sCoefInterceptP}{$p < .001$}
\newcommand{\sCoefPageTypewhitepagesBeta}{-1.26}
\newcommand{\sCoefPageTypewhitepagesZ}{-11.59}
\newcommand{\sCoefPageTypewhitepagesP}{$p < .001$}
\newcommand{\sCoefUsingStructuredOneBeta}{0.96}
\newcommand{\sCoefUsingStructuredOneZ}{10.45}
\newcommand{\sCoefUsingStructuredOneP}{$p < .001$}
\newcommand{\sCoefGroupBBeta}{0.02}
\newcommand{\sCoefGroupBZ}{0.22}
\newcommand{\sCoefGroupBP}{$p = 0.824$}
\newcommand{\sCoefGroupCBeta}{-0.00}
\newcommand{\sCoefGroupCZ}{-0.04}
\newcommand{\sCoefGroupCP}{$p = 0.971$}
\newcommand{\sCoefModeldeepseekVFourFlashBeta}{-0.94}
\newcommand{\sCoefModeldeepseekVFourFlashZ}{-17.26}
\newcommand{\sCoefModeldeepseekVFourFlashP}{$p < .001$}
\newcommand{\sCoefModelgeminiThreeSixFlashBeta}{-0.38}
\newcommand{\sCoefModelgeminiThreeSixFlashZ}{-7.24}
\newcommand{\sCoefModelgeminiThreeSixFlashP}{$p < .001$}
\newcommand{\sCoefModelgptFiveSixSolThinkNoneBeta}{-0.55}
\newcommand{\sCoefModelgptFiveSixSolThinkNoneZ}{-11.02}
\newcommand{\sCoefModelgptFiveSixSolThinkNoneP}{$p < .001$}
\newcommand{\sCoefModelgrokFourFiveBeta}{-0.27}
\newcommand{\sCoefModelgrokFourFiveZ}{-5.42}
\newcommand{\sCoefModelgrokFourFiveP}{$p < .001$}
\newcommand{\sCoefModelkimiKThreeBeta}{-1.06}
\newcommand{\sCoefModelkimiKThreeZ}{-18.55}
\newcommand{\sCoefModelkimiKThreeP}{$p < .001$}
\newcommand{\sCoefPageTypewhitepagesUsingStructuredOneBeta}{0.55}
\newcommand{\sCoefPageTypewhitepagesUsingStructuredOneZ}{4.55}
\newcommand{\sCoefPageTypewhitepagesUsingStructuredOneP}{$p < .001$}

\newcommand{\legacyGoldClaudeLegacy}{54.6\%}
\newcommand{\legacyGoldClaudeOfficial}{54.2\%}
\newcommand{\legacyGoldGptLegacy}{46.1\%}
\newcommand{\legacyGoldGptOfficial}{45.6\%}
\newcommand{\legacyGoldGeminiLegacy}{53.7\%}
\newcommand{\legacyGoldGeminiOfficial}{53.3\%}
\newcommand{\legacyGoldDeepseekLegacy}{35.0\%}
\newcommand{\legacyGoldDeepseekOfficial}{34.5\%}
\newcommand{\legacyGoldGrokLegacy}{55.7\%}
\newcommand{\legacyGoldGrokOfficial}{55.3\%}
\newcommand{\legacyGoldKimiLegacy}{51.2\%}
\newcommand{\legacyGoldKimiOfficial}{50.6\%}
\newcommand{\legacyGoldN}{2,058}
\newcommand{\legacyGoldNModels}{6}
\newcommand{\legacyGoldMeanDelta}{0.44}
\newcommand{\legacyGoldMaxDelta}{0.53}

\newcommand{\ladZeroshot}{42.2\%}
\newcommand{\ladThinking}{47.5\%}
\newcommand{\ladInContext}{63.1\%}


%% file: tables/per_task_overall_typeface_sensitive.tex
\begin{tabular}{lcccccccc}
    \toprule
    \textbf{Task} & $n$ & \textbf{GPT-5.6 Sol} & \shortstack{\textbf{Gemini 3.6} \\ \textbf{Flash}} & \shortstack{\textbf{Grok} \\ \textbf{4.5}} & \shortstack{\textbf{DeepSeek} \\ \textbf{V4 Flash}} & \shortstack{\textbf{Kimi} \\ \textbf{K3}} & \shortstack{\textbf{Claude} \\ \textbf{Opus 5}} & \textbf{Mean} \\
    \midrule
    \multicolumn{9}{l}{\textit{Case Law (open)}} \\
    \midrule
    Basic case citation & 230 & 0.51 & 0.52 & 0.57 & 0.46 & 0.54 & 0.56 & 0.53 \\
    Extended case citation & 74 & 0.28 & 0.30 & 0.31 & 0.20 & 0.34 & 0.34 & 0.30 \\
    Case short form & 65 & 0.75 & 0.88 & 0.83 & 0.15 & 0.85 & 0.83 & 0.72 \\
    \midrule
    \multicolumn{9}{l}{\textit{Enacted Law (open)}} \\
    \midrule
    Federal statutes (print) & 85 & 0.45 & 0.81 & 0.59 & 0.29 & 0.72 & 0.75 & 0.60 \\
    State statutes (print) & 69 & 0.33 & 0.67 & 0.51 & 0.26 & 0.45 & 0.58 & 0.47 \\
    Statutes (electronic) & 38 & 0.11 & 0.05 & 0.05 & 0.05 & 0.08 & 0.05 & 0.07 \\
    Statutes short form & 61 & 0.57 & 0.77 & 0.54 & 0.33 & 0.57 & 0.66 & 0.57 \\
    Administrative law & 42 & 0.40 & 0.43 & 0.48 & 0.19 & 0.40 & 0.45 & 0.39 \\
    \midrule
    \multicolumn{9}{l}{\textit{Other (open)}} \\
    \midrule
    Legislative resources & 43 & 0.33 & 0.63 & 0.47 & 0.56 & 0.37 & 0.60 & 0.49 \\
    Court filings & 57 & 0.19 & 0.12 & 0.16 & 0.04 & 0.16 & 0.42 & 0.18 \\
    Secondary sources & 150 & 0.35 & 0.59 & 0.55 & 0.06 & 0.48 & 0.58 & 0.44 \\
    \midrule
    \multicolumn{9}{l}{\textit{Cloze}} \\
    \midrule
    Case names & 331 & 0.49 & 0.49 & 0.60 & 0.09 & 0.51 & 0.54 & 0.45 \\
    Case reporters & 285 & 0.51 & 0.55 & 0.65 & 0.64 & 0.56 & 0.55 & 0.58 \\
    Case reporters (parallel) & 25 & 0.40 & 0.96 & 0.68 & 0.36 & 0.64 & 0.76 & 0.63 \\
    Case court and date & 305 & 0.54 & 0.54 & 0.63 & 0.65 & 0.55 & 0.57 & 0.58 \\
    Case history & 40 & 0.47 & 0.42 & 0.35 & 0.45 & 0.35 & 0.38 & 0.40 \\
    Case parentheticals & 50 & 0.44 & 0.26 & 0.54 & 0.48 & 0.40 & 0.40 & 0.42 \\
    Signals & 108 & 0.32 & 0.50 & 0.41 & 0.10 & 0.46 & 0.39 & 0.36 \\
    \midrule
    \textbf{Pooled (weighted)} &  & 0.42 & 0.49 & 0.46 & 0.23 & 0.44 & 0.51 &  \\
    \bottomrule
\end{tabular}

%% file: tables/per_task_overall_typeface_insensitive.tex
\begin{tabular}{lcccccccc}
    \toprule
    \textbf{Task} & $n$ & \textbf{GPT-5.6 Sol} & \shortstack{\textbf{Gemini 3.6} \\ \textbf{Flash}} & \shortstack{\textbf{Grok} \\ \textbf{4.5}} & \shortstack{\textbf{DeepSeek} \\ \textbf{V4 Flash}} & \shortstack{\textbf{Kimi} \\ \textbf{K3}} & \shortstack{\textbf{Claude} \\ \textbf{Opus 5}} & \textbf{Mean} \\
    \midrule
    \multicolumn{9}{l}{\textit{Case Law (open)}} \\
    \midrule
    Basic case citation & 230 & 0.71 & 0.71 & 0.75 & 0.65 & 0.73 & 0.77 & 0.72 \\
    Extended case citation & 74 & 0.55 & 0.50 & 0.50 & 0.43 & 0.59 & 0.59 & 0.53 \\
    Case short form & 65 & 0.77 & 0.89 & 0.85 & 0.31 & 0.86 & 0.85 & 0.75 \\
    \midrule
    \multicolumn{9}{l}{\textit{Enacted Law (open)}} \\
    \midrule
    Federal statutes (print) & 85 & 0.66 & 0.86 & 0.87 & 0.64 & 0.82 & 0.81 & 0.78 \\
    State statutes (print) & 69 & 0.58 & 0.67 & 0.61 & 0.49 & 0.62 & 0.64 & 0.60 \\
    Statutes (electronic) & 38 & 0.18 & 0.05 & 0.08 & 0.11 & 0.08 & 0.05 & 0.09 \\
    Statutes short form & 61 & 0.77 & 0.79 & 0.61 & 0.57 & 0.74 & 0.74 & 0.70 \\
    Administrative law & 42 & 0.50 & 0.45 & 0.50 & 0.33 & 0.45 & 0.48 & 0.45 \\
    \midrule
    \multicolumn{9}{l}{\textit{Other (open)}} \\
    \midrule
    Legislative resources & 43 & 0.84 & 0.65 & 0.51 & 0.67 & 0.84 & 0.65 & 0.69 \\
    Court filings & 57 & 0.60 & 0.51 & 0.49 & 0.25 & 0.56 & 0.46 & 0.48 \\
    Secondary sources & 150 & 0.66 & 0.64 & 0.71 & 0.45 & 0.71 & 0.66 & 0.64 \\
    \midrule
    \multicolumn{9}{l}{\textit{Cloze}} \\
    \midrule
    Case names & 331 & 0.74 & 0.68 & 0.81 & 0.49 & 0.77 & 0.83 & 0.72 \\
    Case reporters & 285 & 0.86 & 0.89 & 0.89 & 0.88 & 0.91 & 0.85 & 0.88 \\
    Case reporters (parallel) & 25 & 0.40 & 0.96 & 0.68 & 0.36 & 0.64 & 0.76 & 0.63 \\
    Case court and date & 305 & 0.86 & 0.84 & 0.84 & 0.79 & 0.83 & 0.84 & 0.83 \\
    Case history & 40 & 0.60 & 0.55 & 0.55 & 0.57 & 0.65 & 0.40 & 0.55 \\
    Case parentheticals & 50 & 0.70 & 0.54 & 0.84 & 0.72 & 0.72 & 0.80 & 0.72 \\
    Signals & 108 & 0.36 & 0.63 & 0.56 & 0.16 & 0.68 & 0.44 & 0.47 \\
    \midrule
    \textbf{Pooled (weighted)} &  & 0.67 & 0.63 & 0.62 & 0.44 & 0.70 & 0.62 &  \\
    \bottomrule
\end{tabular}